\documentclass[preprint,12pt,authoryear]{elsarticle}

\usepackage{amssymb}
\usepackage{amsmath}
\usepackage{multirow}
\usepackage{array}
\usepackage{booktabs}
\usepackage{rotating}

\usepackage{hyperref} 
\hypersetup{pdfborder = {0 0 0}}
\usepackage[acronym]{glossaries}
\newacronym{2ch}{2CH}{two-chamber long-axis}
\newacronym{4ch}{4CH}{four-chamber long-axis}
\newacronym{acdc}{ACDC}{Automated Cardiac Diagnosis Challenge}
\newacronym{afd}{aFD}{\textit{average Frame Difference}}
\newacronym{aha}{AHA}{American Heart Association}
\newacronym{arr}{ARR}{Congenital Arrhythmogenesis} 
\newacronym{arv}{ARV}{Abnormal right ventricle} 
\newacronym{cfd}{$cFD$}{cyclic frame difference}
\newacronym{cia}{CIA}{Interatrial communication} 
\newacronym{cmr}{CMR}{cardiovascular magnetic resonance}
\newacronym{cnn}{CNN}{Convolutional Neuronal Network}
\newacronym{cvd}{CVD}{Cardiovascular diseases}
\newacronym{dcm}{DCM}{Dilated cardiomyopathy }
\newacronym{dlv}{DLV}{Dilated left ventricle} 
\newacronym{drv}{DRV}{Dilated right ventricle}
\newacronym{ecg}{ECG}{electrocardiogram}
\newacronym{ed}{ED}{end-diastole}
\newacronym{es}{ES}{end-systole}
\newacronym{gcn}{GCN}{German Competence Network}
\newacronym{hcm}{HCM}{Hypertrophic cardiomyopathy}
\newacronym{iov}{IOV}{inter-observer variability}
\newacronym{iqr}{IQR}{interquartile range}
\newacronym{lax}{LAX}{long-axis}
\newacronym{lstm}{LSTM}{long short-term memory}
\newacronym{lv}{LV}{left ventricle}
\newacronym{mm2}{M\&Ms-2}{ Multi-Centre, Multi-View, Multi-Vendor \& Multi-Disease Cardiac Image Segmentation Challenge}
\newacronym{mnms}{M\&Ms}{Multi-Centre, Multi-Vendor \& Multi-Disease Cardiac Image Segmentation Challenge}
\newacronym{md}{MD}{mid diastole}
\newacronym{miccai}{MICCAI}{Medical Image Computing and Computer Assisted Intervention}
\newacronym{minf}{MINF}{Myocardial infarction}
\newacronym{mri}{MRI}{magnetic resonance imaging}
\newacronym{ms}{MS}{mid systole}
\newacronym{nor}{NOR}{Healthy} 
\newacronym{pf}{PF}{peak flow}
\newacronym{rnn}{RNN}{Recurrent Neural Network}
\newacronym{rv}{RV}{right ventricle}
\newacronym{sax}{SAX}{short-axis}
\newacronym{sop}{SOP}{standard operating procedure}
\newacronym{ssim}{SSIM}{structural similarity index measure}
\newacronym{stacom}{STACOM}{Statistical Atlases and Computational Modelling of the Heart}
\newacronym{tri}{TRI}{Tricuspidal Regurgitation}
\newacronym{tof}{TOF}{Tetralogy of Fallot}

\journal{Medical Imaga Analysis}

\begin{document}

\begin{frontmatter}



\title{Deformable Image Registration for Self-supervised Cardiac Phase Detection in Multi-View Multi-Disease Cardiac Magnetic Resonance Images} %% Article title


%% Author names
\author[UKHD]{Koehler, Sven\corref{cor1}} 
\author[UKHD,UMHD]{Mueller, Sarah Kaye\corref{cor1}}
\author[UKHD,DZHK]{Kiekenap, Jonathan}
\author[DALLAS]{Greil, Gerald}
\author[DALLAS]{Hussain, Tarique}
\author[DZHK,GCN,HANNOVER]{Sarikouch, Samir}
\author[UKHD]{Andre, Florian}
\author[UKHD]{Frey, Norbert}
\author[UKHD,UMHD,DZHK]{Engelhardt, Sandy}
\cortext[cor1]{Joint first authorship}
%Department and Organization

\affiliation[UKHD]{
            organisation = {Department of Internal Medicine III, Heidelberg University Hospital},
            city = {Heidelberg},
            postcode={69120}, 
            country = {Germany}}

\affiliation[UMHD]{
            organisation = {Medical Faculty of Heidelberg University},
            city = {Heidelberg},
            postcode={69120}, 
            country = {Germany}}

\affiliation[DZHK]{
            organisation={DZHK (German Centre for Cardiovascular Research)},
            city={Heidelberg},
            postcode={69120}, 
            % state={Baden-Würtemberg},
            country={Germany}}

\affiliation[DALLAS]{
            organisation={Division of Pediatric Cardiology, Department of Pediatrics, UT Southwestern /Children’s Health},
            city = {Dallas},
            country={USA}}

\affiliation[GCN]{
            organisation={German Competence Network for Congenital Heart Defects},
            city = {Berlin},
            country={Germany}}
            
\affiliation[HANNOVER]{
            organisation={Department of Cardiothoracic, Transplantation and Vascular Surgery, Hannover Medical School},
            city = {Hannover},
            country={Germany}}

%% Abstract
\begin{abstract}
%% Text of abstract max 250 words.
Cardiovascular magnetic resonance (CMR) is the gold standard for assessing cardiac function, but individual cardiac cycles complicate automatic temporal comparison or sub-phase analysis. Accurate cardiac keyframe detection can eliminate this problem. However, automatic methods solely derive end-systole (ES) and end-diastole (ED) frames from left ventricular volume curves, which do not provide a deeper insight into myocardial motion.

We propose a self-supervised deep learning method detecting five keyframes in short-axis (SAX) and four-chamber long-axis (4CH) cine CMR. Initially, dense deformable registration fields are derived from the images and used to compute a 1D motion descriptor, which provides valuable insights into global cardiac contraction and relaxation patterns. From these characteristic curves, keyframes are determined using a simple set of rules.

The method was independently evaluated for both views using three public, multicentre, multidisease datasets. M\&Ms-2 (n=360) dataset was used for training and evaluation, and M\&Ms (n=345) and ACDC (n=100) datasets for repeatability control. Furthermore, generalisability to patients with rare congenital heart defects was tested using the German Competence Network (GCN) dataset.

Our self-supervised approach achieved improved detection accuracy by 30\% - 51\% for SAX and 11\% - 47\% for 4CH  in ED and ES, as measured by cyclic frame difference (cFD), compared with the volume-based approach.  
We can detect ED and ES, as well as three additional keyframes throughout the cardiac cycle with a mean cFD below 1.31 frames for SAX and 1.73 for LAX. Our approach enables temporally aligned inter- and intra-patient analysis of cardiac dynamics, irrespective of cycle or phase lengths. GitHub repository: \url{https://github.com/Cardio-AI/cmr-multi-view-phase-detection.git}

\end{abstract}

% \includegraphics[width=\textwidth]{Figure_0_Visual_Abstract.pdf}

%%Research highlights: 3 to 5 bullet points, each a maximum of 85 characters, including spaces
% \begin{highlights}
% \item Generation of one-dimensional time resolved global cardiac motion-curves from high-definition, 2D+t and 3D+t \acrshort{cmr} , short-axis and long-axis view
% \item Requires no manual annotations while training in a self-supervised manner.
% \item Disease and scanner independent global cardiac motion curves for temporal alignment of scans from patients
% \item Enable the robust detection of five different cardiac keyframes throughout the cardiac cycle
% \item Keyframe detection has a mean cyclic frame difference of less than 1.31 frames for short-axis and 1.58 for four-chamber long-axis view

% \end{highlights}

%% Keywords 1 to 7
\begin{keyword}
Cardiac Phase Detection \sep Cardiac Motion Description \sep Cardiac Magnetic Resonance Imaging \sep Self-supervised Learning \sep Discrete Vector Fields 

% %% MSC codes here, in the form: \MSC code \sep code
% %% or \MSC[2008] code \sep code (2000 is the default)
\end{keyword}

\end{frontmatter}


\begin{figure}[h]
    \centering
    \includegraphics[width=\textwidth]{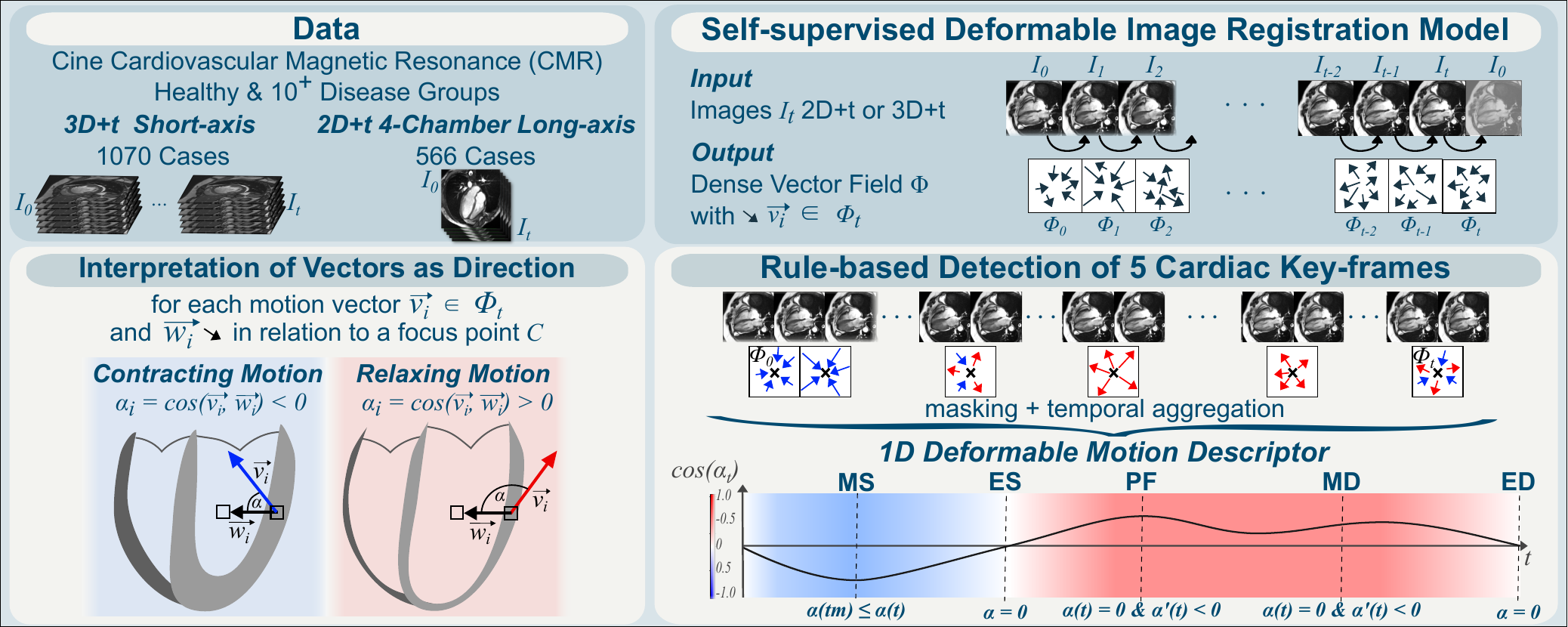}
    \caption{\textbf{Graphical abstract.} Overview of the proposed pipeline. The top row illustrates the input data and the self-supervised deformable image registration model. The bottom row shows the interpretation of the resulting dense deformable vector field as motion direction, enabling derivation of a one-dimensional motion descriptor for cardiac key-frame detection.}
    \label{fig:placeholder}
\end{figure}

\section{Introduction}
\label{sec1:Introduction} 
\acrfull{cvd} were responsible for 17.9 million deaths worldwide in 2019 \citep{who_cvd_deaths19}. In the context of the diagnosis of \acrshort{cvd}s, \acrfull{cmr} imaging is widely regarded the gold standard for detailed cardiac evaluation. This is primarily due to its capacity to capture the dynamic processes of the heart and provide high-contrast soft tissue images. However, reliable inter- and intra-patient comparisons of \acrshort{cmr} images are often hindered by temporal misalignment due to physiological variations in heartbeats and differences in imaging protocols and resolutions.  Therefore, it is essential to define cardiac keyframes in  \acrshort{cmr} for alignment and interpolation to facilitate more accurate comparison. 

The cardiac cycle is divided into two main phases: Diastole, consisting of iso-volumetric relaxation followed by filling of the ventricles first by passive chamber enlargement and second by atrial contraction, and systole, including iso-volumetric contraction and ejection of blood into the body and lungs. Several parameters are instrumental in the evaluation of cardiac morphology, diagnosis of \acrshort{cvd}s and clinical decision making, including cardiac chamber size, wall thickness, global and peak systolic strain, ejection fraction, and stroke volume \citep{mada_2014_10_010}. They are measured during and between \acrfull{ed} and \acrfull{es}, phases which are of particular interest.

Traditionally, the detection of cardiac keyframes relies on manual annotation. This approach is not only time-consuming, but also susceptible to observer bias, leading to a median inter-observer variability of three frames for \acrshort{ed} and \acrshort{es} \citep{zolgharni_2D_echo}. The use of the QRS complex derived from \acrfull{ecg} signals has been shown to constitute an effective approach for automatic identification of \acrshort{ed}. Nevertheless, even when \acrshort{ecg} signals are available for analysis, they are frequently not consistently retained with \acrshort{cmr} imaging data, and most of the time distorted by the magnetic field of the magnetic resonance scanner. This limits the applicability of this approach.

In this study, we demonstrate a method for generation of one-dimensional motion descriptor that reflect systolic and diastolic motion patterns of the cardiac cycle from \acrfull{4ch} and \acrfull{sax} \acrshort{cmr} images in a self-supervised manner.  The generation of this descriptor is achieved through the utilisation of deformable image registration fields, for which we employ 2D and 3D U-net models. The method facilitates the identification of five keyframes including \acrshort{es} and \acrshort{ed} in multi-view \acrshort{cmr}s. It offers a fully automatic solution that does not require external labels or \acrfull{ecg} data.

\subsection{Related Work}
\label{sec2:relatedWork}
The field of cardiac image analysis has witnessed significant advancements in the domain of deep learning, particularly in the areas of segmentation, registration, and regression. This section presents a review of studies that have focused on the registration and analysis of cardiac motion, as well as those that have addressed the problem of cardiac keyframe detection.

\subsubsection{Cardiac Keyframe Detection}
Automatic methods operating independently of associated \acrshort{ecg} signals  for cardiac keyframe detection have been widely explored in echocardiography \citep{kachenoura2007automatic, barcaro_echo, gifani2010automatic, darvishi_echo, shalbaf2015echocardiography,  dezaki2018cardiac, fiorito2018detection, LANE_echo2021104373} and, to a lesser extent, in \acrshort{cmr} \citep{phase_detection_kong2016, yang2017convolutional, xue2018cmr, garcia2023cardiac}. The early approaches range from semi-automatic approaches that require manual input \citep{kachenoura2007automatic, barcaro_echo, darvishi_echo} to more advanced techniques utilising non-linear dimensionality reduction techniques \citep{gifani2010automatic, shalbaf2015echocardiography}. However, the latter ones have only been evaluated on small patient cohorts ($n = 8$ and $n = 32$).

Recent deep learning developments have shown promising potential in capturing spatial and temporal features for more robust cardiac keyframe detection.
\cite{fiorito2018detection} utilised a 3D \acrfull{cnn} to extract spatio-temporal features and joined it with a \acrfull{lstm} to classify between diastolic and systolic frames, where  \acrshort{ed} and \acrshort{es} where automatically identified as the transition between both states.  Their approach achieved an \acrfull{afd} of 1.52/1.48 ( \acrshort{ed}/\acrshort{es}).
The supervised approach of \cite{dezaki2018cardiac} achieved an even more precise detection of  \acrshort{ed} with an aFD of 0.71/1.92 ( \acrshort{ed}/\acrshort{es}), by introducing a Densely Gated \acrfull{rnn}, which uses temporal dependencies and a global extrema loss function.
The best results achieved \cite{LANE_echo2021104373} with an average absolute frame difference of 0.66/0.81 ( \acrshort{ed}/\acrshort{es}). They framed  \acrshort{ed} and \acrshort{es} detection as a regression problem using \acrshort{cnn}s, trained and tested on multi-centre datasets.

Although echocardiography has been extensively researched in the context of cardiac phase detection, comparatively little research has been dedicated to \acrshort{cmr}.
To address this gap,  \cite{phase_detection_kong2016} proposed a hybrid \acrshort{rnn}-\acrshort{cnn} architecture for cardiac keyframe detection in \acrshort{cmr} with impressive average frame difference (aFD) of 0.38/0.44 ( \acrshort{ed}/\acrshort{es}). However, their approach was restricted to a homogeneous single-centre private dataset with uniform sequence lengths starting with  \acrshort{ed}, which restricts its broader applicability.

The architecture of \cite{xue2018cmr} employs multiple \acrshort{rnn}s and \acrshort{cnn}s to quantify the \acrfull{lv} dimensions and identify the cardiac keyframes of a private multi-centre, multi-pathology dataset comprising single slice short-axis \acrshort{cmr} images with identical sequence length. They classified each frame into diastole or systole, achieving an error rate of  8.2\%.

\begin{figure*}
    \centering
    \includegraphics[width=0.8\linewidth]{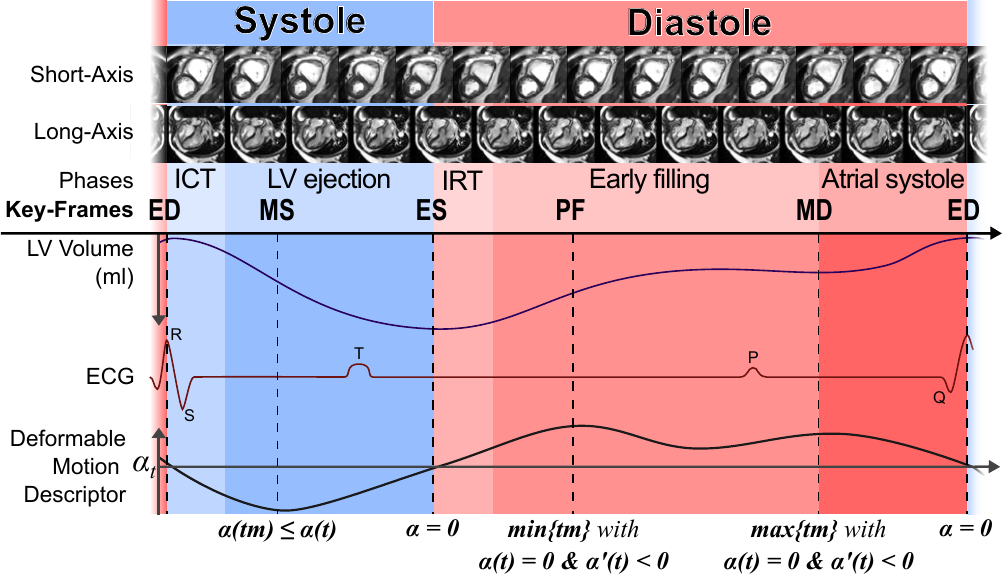}
    \caption{\textbf{LV volume curve, ECG signal and our proposed motion descriptor $\alpha$ over the cardiac cycle.}  
    The figure shows the temporal relation between cardiac phases and keyframes, left ventricular volume (top blue curve), \acrshort{ecg} (middle red curve), and the motion descriptor $\alpha$ (bottom black curve) derived from \acrshort{cmr} data. The cycle is divided into systole (blue) and diastole (red), with iso-volumetric contraction time (ICT) and iso-volumetric relaxation time (IRT) in respectively lighter shades. The bottom curve depicts $\alpha$, where a negative value for $\alpha$ indicates contractile motion, and positive values refer to relaxing cardiac motion.  Characteristic points in $\alpha$ align with physiological events (Section \ref{sec2_3:cardiac_key_frames}).}
    \label{fig:cardiaC_curves}
\end{figure*}

The aforementioned approaches in \acrshort{cmr} predominantly rely on private and single-modality datasets, which limit their generalisability. To overcome this,  \cite{garcia2023cardiac} developed an architecture for the detection of  \acrshort{ed} and \acrshort{es} frames from \acrshort{sax} \acrshort{cmr} images utilizing the publicly available \acrshort{mm2} dataset. Their approach integrates a pre-trained \acrshort{cnn} for segmentation with a sequential module, either consisting of a LSTM or a Transformer encoder, to detect cardiac keyframes. Their LSTM-architecture achieved the best results with an \acrfull{afd} of 1.70/1.75 (\acrshort{ed}/\acrshort{es}).
An overview of the different approaches can be found in \ref{tbl:related_work_key-frame}

Despite promising results, the aforementioned methods in \acrshort{cmr} face significant limitations. As most methods are trained on homogeneous datasets collected from single centres, their ability to generalise to unseen data is restricted, especially when confronted with scans from different scanner types or rare cardiac conditions. Moreover, their reliance on labelled data for supervised learning makes the adaption to new scenarios cumbersome, as retraining would require time-expansive relabelling.  In addition, many approaches are based on the assumption that alterations in \acrshort{lv} volume can be used for the detection of cardiac phases. Nevertheless, this assumption does not universally hold true as iso-volumetric contraction and relaxation, which occur near  \acrshort{ed} and \acrshort{es}, exhibit myocardial changes without corresponding shifts in ventricular volume as shown in the schematic illustration presented in Fig. \ref{fig:cardiaC_curves}.

\subsubsection{Cardiac Image Registration}
In recent years, significant advancements have been made in the field of image registration, leading to the introduction of numerous methods for the accurate quantification of myocardial deformation from cine \acrshort{cmr} images. Key contributions in this field utilize \acrshort{cnn} architectures, such as  \cite{cmr_motion_estimation_Qin_2018,  Dalca_2019, cmr_probabilistiC_diffeomorphiC_registration_Krebs_2019, cmr_probabilistiC_registration_Krebs_2020, cmr_motion_MulViMotion_Meng_2022}.

In their work, \cite{cmr_motion_estimation_Qin_2018} put forth a network comprising two branches, which share a joint multi-scale feature encoder. The first branch is responsible for estimating motion, which is achieved through the use of an unsupervised Siamese-style recurrent spatial transformer network. The second branch performs a segmentation, accomplished through the deployment of a \acrlong{cnn}. The framework developed by \cite{Dalca_2019}, Voxelmorph, consists of a probabilistic generative model with an inference algorithm based on unsupervised learning. While their framework enforces a multivariate Gaussian distribution for each component of the velocity field to measure uncertainty, it does not learn global latent variable models.

The self-supervised probabilistic motion model as proposed by \cite{cmr_probabilistiC_diffeomorphiC_registration_Krebs_2019} employs a learning process to identify the deformation model from a set of training images. This approach focuses on reconstructing a fixed image  \(I_t\) from the moving image \(I_0\). Potential applications include the simulation of pathologies or the completion of missing sequences. The model comprises an encoder for mapping images to a latent space, a Temporal Convolutional Network (TCN) for temporal modeling, and a decoder to generate deformation fields. These deformation fields are used to warp the moving image and reconstruct the fixed image. The model was trained on short-axis \acrshort{cmr} sequences from the \acrshort{acdc} challenge \citep{acdc_dataset_8360453}. 

The aforementioned methods \citep{cmr_motion_estimation_Qin_2018, Dalca_2019, cmr_probabilistiC_diffeomorphiC_registration_Krebs_2019, cmr_probabilistiC_registration_Krebs_2020} are only capable of registering in-plane motion for individual \acrshort{cmr} slices, rendering them unsuitable for 3D \acrshort{sax} registration. This limitation is further amplified by slice misalignment, which introduces additional complexity to through-plane motion registration. To address this issue, \cite{cmr_motion_MulViMotion_Meng_2022} presented the multi-view motion estimation network (MulViMotion). This employs a hybrid 2D/3D architecture, comprising a FeatureNet (2D \acrshort{cnn}s) and a MotionNet (3D \acrshort{cnn}s). This combination enables the model to register both the in-plane and through-plane motion. The study utilised data from 580 subjects with both \acrshort{sax} and \acrshort{4ch} views from the UK Biobank study. Additionally, their study relies on ground truth labels for accurate motion estimation.

\subsection{Contributions}

This work introduces a fully self-supervised architecture as base for a robust one-dimensional motion descriptor that captures the contraction and relaxation patterns of the cardiac cycle. The hypothesis underpinning this work is that cardiac keyframe detection based on myocardial displacement fields is more accurate than using \acrshort{lv} volume change. The architecture is capable to detect traditional  \acrshort{ed} and \acrshort{es} frames, as well as three additional keyframes, independent of sequence length. 

This enables reliable temporal alignment for inter- and intra-patient comparisons of cardiac function across the cardiac cycle and can be used for delineation of different disease cohorts. This was previously investigated in our recent work of \cite{koehler2025strain}. In that work, the identification of additional keyframes facilitated aligned strain calculation, thereby achieving a more significant diagnostic value in the detection of scarred and fibrotic tissue than the conventional approach in patients with Duchenne muscular dystrophy.

Building on our previous approach for self-supervised keyframe detection \citep{koehler2022self}, we refine the post-processing for improved keyframe detection in \acrshort{sax}. We also extend the method to handle both 3D stacks of cine SSFP \acrshort{sax} and 2D \acrshort{4ch} \acrshort{cmr} images. 
Moreover, we conduct a much more comprehensive series of experiments to evaluate the performance and comparison on a range of datasets, thereby demonstrating the significant advantages of the proposed keyframe detection method in comparison to existing state-of-the-art techniques and inter-observer variability. Our method addresses generalisation by being independent of labelled data, and demonstrate robust performance across multi-centre, multi-pathology, multi-scanner, and multi-view \acrshort{cmr} datasets, ensuring its applicability in diverse clinical settings, including rare diseases.

% The remainder of the paper is structured as follows. Section \ref{sec2:relatedWork} provides an overview of related work on deformable image registration and cardiac keyframe detection. Section \ref{sec3:methods} presents our proposed method in detail. Section \ref{sec4:ExperimentResults} presents the experimental results, with a particular focus on cardiac keyframe detection. Finally, Section \ref{sec5:discussion} discusses the results and Section \ref{sec6:Conclusion} presents the conclusions. 

\section{Material and Methods}
\label{sec3:methods}
This work is based on the premise that sequential deformable registration fields can effectively capture the dynamic nature of the heart. While 3D+t deformable dense vector fields \(\phi_t\) provide a detailed representation of cardiac motion, they present key limitations, including high dimensionality and the inclusion of non-cardiac tissue deformations. To address these challenges, we propose a compact 1D motion descriptor $\alpha_t$ that captures the essential cardiac contraction and relaxation pattern over time, after automatic filtering of most non-cardiac related structures. Derived from \(\phi_t\), this scalar signal encodes directional motion relative to a fixed reference point, making the motion description independent of the image grid.

Our approach comprises three modules: (1) a deformable registration model that estimates cardiac motion as a discrete vector field \(\phi_t\), with each motion vector \(\overrightarrow{v} \in \phi_t\) (Section \ref{sec2_1:registration}); (2) a motion descriptor module that computes $\alpha_t$ by masking and aggregating \(\phi_t\) in relation to a focus point $C$, along with the corresponding norm curve \(|\overrightarrow{v}|_t\)(Section \ref{sec2_2:motion}); and (3) a rule-based module that detects the cardiac keyframes from $\alpha_t$ (Section \ref{sec2_3:cardiac_key_frames}). These components are described in detail below along with the datasets, evaluation metrics, and experimental setup.

\subsection{Deformable Registration Model}
\label{sec2_1:registration}
Due to the varying spatial dimensions registration models have to be trained separately for each view. The image sequence is defined as $I$, where $I_t$ represents either the 3D image stack of cine SSFP \acrshort{sax} \acrshort{cmr} or a 2D single slice \acrshort{4ch} \acrshort{cmr} image at time point $t=[1,\dots,T]$. 
The deformable image registration task is defined as \(\phi,\hat{M}=f_\Theta(M,F)\) in the spatial domain \(\mathbb{R}^2\) for \acrshort{4ch} images and \(\mathbb{R}^3\) for \acrshort{sax} volumes. Here, \(M\) and \(F\) represent the moving and fixed image pairs from the same \acrshort{cmr} sequence, where \(M=I_t\) and \(F=I_{t+1}\). The function \(f\), parametrized by learnable weights \(\Theta\),  generates the resulting discrete vector field \(\phi\) and the moved image \(\hat{M}\). The moved image \(\hat{M}\) is obtained by applying \(\phi\)  to \(M\) using a spatial transformer layer as proposed by \cite{Jadeberger_2015_33ceb07b}. 

Since the target for interpolation is the previous frame \(I_{t-1}\), the resulting discrete vector field \(\phi\)  is a forward displacement field, commonly referred to as pull-registration. This registration field between two sequential cine \acrshort{cmr} frames can be interpreted as the sequential motion or displacement field of each voxel throughout the cardiac cycle.

The registration loss, as defined by Equation \ref{eq:loss}, consists of two components: an image similarity component $\mathcal{L}_{sim}$ and a regularisation term $\mathcal{L}_{smooth}$. 
For $\mathcal{L}_{sim}$, we employ the \acrfull{ssim}, which has demonstrated superior performance in our previous work \citep{koehler2022self}. The 2D \acrshort{ssim}, as shown in Equation \ref{eq:ssim}, quantifies the resemblance between two images based on their luminance, contrast and structure. In our case, the images annotated as $I_t$ and $I_{t+1}$ represent two consecutive time steps. For the 3D \acrshort{sax} model, we average the 2D \acrshort{ssim} values across each 3D volume. 
% Here, $\mu_x, \mu_y$ represent the average and $\sigma_x^2, \sigma_y^2$ signify the variances of a NxN region in x and y. The co-variance of $I$ and $y$ is given by $\sigma_{xy}$ and the two variables $C_1, c2$ are included to avoid instability. 
The regularisation term, $\mathcal{L}_{smooth}$ (Equation \ref{eq:smooth}), is based on a diffusion regulariser, as described by \cite{Balakrishnan_2018_CVPR}. This regulariser enforces smoothness in the spatial gradients of the deformation field $\phi$ over the voxel space $\Omega$ in $I_t$. The regularisation parameter $\lambda$ was set to 0.001.

\begin{equation}
    \mathcal{L}(F, M, \phi) = \mathcal{L}_{sim}(F, M(\phi)) + \lambda\mathcal{L}_{smooth}(\phi) 
    \label{eq:loss}
\end{equation}

\begin{equation}
    SSIM(I_t, I_{t+1}) = \frac{(2\mu_{I_t}\mu_{I_{t+1}}+ C_1)(2\sigma_{I_tI_{t+1}}+C_2)}{(\mu_{I_t}^2+\mu_{I_{t+1}}^2+ C_1)(\sigma_{I_t}^2+\sigma_{I_{t+1}}^2+C_2)}
    \label{eq:ssim}
\end{equation}

\begin{equation}
    \mathcal{L}_{smooth}(\phi) = \sum_{p\in\Omega}||\triangledown\phi(p)||^2
    \label{eq:smooth}
\end{equation}

Due to the different dimensions of both views, we employ a 3D CNN-based sequential volume-to-volume deformable registration module for \acrshort{sax} \acrshort{cmr}. It consists of a modified time-distributed 3D U-Net architecture, inspired by \cite{Ronneberger_2015_U-net}, followed by a spatial transformer layer, similar to \cite{Balakrishnan_2018_CVPR}. 
The input to our final \acrshort{sax} model is a 4D volume with dimensions b × 40 × 16 × 64 × 64, representing batch size, time, spatial slices, and x/y dimensions, respectively. 

Given the 2D+t nature of the \acrshort{4ch} sequences, we utilize a deformable registration module based on 2D CNN, using a U-Net architecture. The input to this model is a 3D volume with dimensions similar to those of the \acrshort{sax} model, but without the dimension for spatial slices and a higher in-plane resolution, resulting in the input layer with the dimensions b × 40 × 288 × 288. For further details, please refer to our GitHub repository\footnote{\url{https://github.com/Cardio-AI/cmr-multi-view-phase-detection.git}}.

The direction module (Section \ref{sec2_2:motion}) processes the output displacement field to compute voxel-/pixel-wise $\alpha_i$ and $|\overrightarrow{v_i}|$  values. From these spatial maps, the one-dimensional motion descriptor $\alpha_t$ and its magnitude curve $|\overrightarrow{v}|_t$ are derived per 4D/3D volume, which are utilized in our rule-based framework (Section \ref{sec2_3:cardiac_key_frames}).
Importantly, all components of our model are differentiable, enabling end-to-end learning in a supervised setting.

\subsection{One-Dimensional Motion Descriptor}
\label{sec2_2:motion}

To compactly represent the global direction of cardiac motion over time, we derive a one-dimensional temporal descriptor $\alpha_t$ from the dense deformation field $\phi_t$. This descriptor aggregates the directional information of voxel-wise displacements, masked to restrict the analysis to regions of relevant cardiac motion: 
\begin{equation}
\alpha_t = \mathcal{A}\left( \left\{ M(\mathbf{x}_i) \cdot \phi_t(\mathbf{x}_i) \right\}_{i=1}^N \right),    
\end{equation}

where $M(\mathbf{x}_i) \in \{0,1\} $ is a binary spatial mask and $\mathcal{A}$ denotes a directional aggregation operator, which summarizes the dominant motion direction relative to a defined focus point $C$. Here $\phi_t(\mathbf{x}_i)$ is the voxel-wise displacement vector at time t, from which the directional descriptor $\alpha_i$ is computed as described below. The operator $\mathcal{A}$ thus implicitly acts on the directional quantities derived from $\phi_t$.

\begin{figure*}[ht!]
    \centering
    \includegraphics[width=0.75\linewidth]{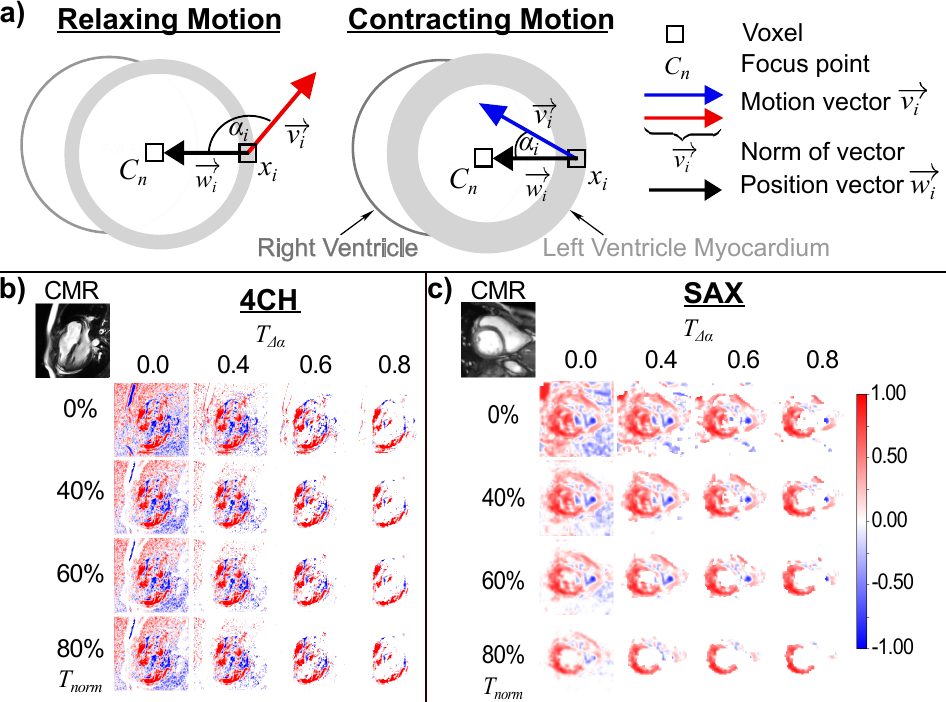}
    \caption{\textbf{Self-supervised rule-based masking of \acrshort{cmr}} \textbf{a)} Schematic illustration of computation of the direction of motion $\alpha$. The motion vector $\vec{v}$ from each voxel $\mathbf{x}_i$ is compared to a reference position vector $\vec{w}$, which points from the corresponding voxel to a fixed anatomical focus point $C_n$. The angle between $\vec{v}$ and $\vec{w}$ is quantified by their cosine similarity $\alpha = \cos(\vec{v}, \vec{w}) \in [-1,1]$. This scalar $\alpha$ represents the directional relationship: negative values below $0$ indicate contraction (motion toward $C$), while positive values indicate relaxation (motion away from $C_n$). 
    \textbf{b \& c)} The first row shows the original \acrshort{cmr} slice at a single time point from either the \acrshort{4ch} (b) or \acrshort{sax} (c) view. The grid below presents filtered directional motion fields $\alpha$ for the same frame, visualized at varying thresholds. Columns correspond to increasing directional change thresholds $T_{\Delta\alpha}$, and rows to increasing motion magnitude percentiles $T_{\text{norm}}$. Blue indicates contractile motion ($-1 \leq \alpha < 0$) directed toward the focus point $C$, and red indicates relaxing motion  ($0 < \alpha \leq 1$) away from it. The top-left cell ($T_{\Delta\alpha} = 0.0$, $T_{\text{norm}} = 0$) shows the raw, unfiltered deformation field. %The masking procedure progressively filters out non-cardiac or noisy motion using a norm-based and direction-based thresholding strategy to isolate regions of consistent cardiac contraction or relaxation.
    }
    \label{fig:masking_example}
\end{figure*}

Let the voxel space $\Omega$ in $I_t$ be defined over \( N \) voxels/pixels \(\Omega= \{\mathbf{x}_i\}_{i=1}^N \), where \(x_i \in \mathbb{Z}^d\) and $d \in \{2,3\}$ depends on the \acrshort{cmr} image view, 2D for \acrshort{4ch} and 3D \acrshort{sax}, respectively. At each location $x_i$ and time \(t\), the displacement vector is given by \(\overrightarrow{v_i} = \phi_t(\mathbf{x}_i) \in \mathbb{R}^d \).  The positional reference vector $\overrightarrow{w_i} = C - \mathbf{x}_i \in \mathbb{R}^d$ points from each voxel to the focus point \(C \in \mathbb{Z}^d\). The focus point $C$ can be represented by an anatomical landmark, if segmentation information is available, or computed without prior knowledge in an unsupervised manner (see Section \ref{chapter2_6:experimental_setup} for further details). 

The direction of motion for each spatial location $\mathbf{x}_i$ is quantified by cosine similarity between the displacement and the position vector:
\begin{align}
    & \alpha_i = cos(\overrightarrow{v_i}, \overrightarrow{w_i}) = \alpha_i = \frac{\overrightarrow{v_i} \cdot \overrightarrow{w_i}}{\|\overrightarrow{v_i}\| \, \|\overrightarrow{w_i}\|}, \quad \alpha_i \in [-1, 1].
 % & \text{ with } \alpha_i\in[-1,1].
\end{align}

where \(\alpha \in [-1, 1]\) indicates the direction of the motion. Hereby, $\alpha < 0$ is interpreted as contractile motion (towards $C$), and \(\alpha >0\) indicates relaxing motion (away from $C$),  as illustrated in Figure \ref{fig:masking_example}a).

The global motion descriptor $\alpha_t$ is computed by averaging over the masked region:
\begin{equation}
 \alpha_t = \frac{1}{\sum_{i=1}^N M(\mathbf{x}_i)} * \sum_{i=1}^N M(\mathbf{x}_i) a_i.
% \alpha_t = \overline{M(\mathbf{x}_i)\cdot a_i} = \frac{1}{\sum_{i=1}^N M(\mathbf{x}_i)} * \sum_{i=1}^N M(\mathbf{x}_i) \cdot cos(\overrightarrow{v_i}, \overrightarrow{w_i}).
\end{equation}

The resulting motion descriptor $\alpha_t$ is smoothed with the Gaussian filter (\(\sigma = 2\)) and subsequently used for keyframe detection (Section \ref{sec2_3:cardiac_key_frames}).
The smoothing of \(\alpha_t\) may lead to minor shifts in the zero-crossing time points ($<1$ frames), corresponding to $t_{ \acrshort{ed}}$ and $t_{\acrshort{es}}$ but also eliminates spurious zero crossings associated with weak or pathological relaxation phases.

The masking is performed according to predefined rules, though anatomical knowledge (e.g. segmentation masks) can also guide the region of interest for anatomical mapping of the motion descriptor.
To robustly exclude non-cardiac motion and noise, we construct a rule-based binary mask $M:\Omega\xrightarrow{}\{0,1\}$, defined as:

\begin{align}
   & M(\mathbf{x}_i)= M_{||\cdot||}(\mathbf{x}_i) \cdot M_{\Delta\alpha}(\mathbf{x}_i), & \forall\mathbf{x}_i \in \Omega 
\end{align}

Here, $ M_{||\cdot||}(\mathbf{x}_i)$ filters based on displacement magnitude of the motion vector $|\overrightarrow{v}|$, and  $M_{\Delta\alpha}(\mathbf{x}_i)$ based on temporal directional change.

The magnitude filter $M_{||\cdot||}(\mathbf{x}_i)$ retains $x_i$ with sufficiently large displacement magnitude across time. Let the temporally averaged displacement magnitude be defined as:
\begin{equation}
\|\overrightarrow{v}_i\| = \frac{1}{T} \sum_{t=1}^{T} \|\phi_t(\mathbf{x}_i)\|    .
\end{equation}

The binary magnitude filter is then given by:
\begin{equation}
\begin{aligned}
    M_{||\cdot||}(\mathbf{x}_i) = H(||\overrightarrow{v}_i|| - T_{norm}),  \quad T_{norm} \in [0,100],
\end{aligned}    
\end{equation}
where $H(\cdot)$ is the Heaviside function, and $T_{norm}$ is a chosen percentile of the magnitude distribution (e.g., 40th, 50th, 60th). 

Next, $M_{\Delta\alpha}(\mathbf{x}_i)$ is applied to retain only $\mathbf{x}_i$ with minimal directional change $\Delta\alpha$ over time:

\begin{align}
M_{\Delta\alpha}(\mathbf{x}_i) = H(\Delta \alpha_i - T{\Delta\alpha}) \\
\Delta \alpha_i = \max_t\left(\alpha_i(t)\right) - \min_t\left(\alpha_i(t)\right),
\end{align}
Here,  $\Delta\alpha$ is the discrepancy between the maximum and minimum values of $\alpha$ at each voxel $\mathbf{x}_i$ over the entire sequence, designated as $\max_t\left(\alpha_i(t)\right)$  and $\min_t\left(\alpha_i(t)\right)$ respectively. This mask is designed to eliminate noisy or non-directional motion from the area of interest. Voxels exhibiting minimal directional change, such as static noise or predominantly unidirectional flow in vessels, demonstrate negligible  variation in motion direction in relation to a focus point inside the heart. In contrast, myocardial voxels demonstrate clear pulsatile motion relative to the cardiac focus point.
Since $\alpha_t\in [-1,1]$, both $T_{\Delta\alpha}$ and $\Delta\alpha$ fall within the range $[0,2]$.

The optimal threshold value for the displacement magnitude and the minimal directional change were identified empirically. Examples of the resulting masked direction fields under different thresholds for a single \acrshort{4ch} and \acrshort{sax} slice are shown in Figure \ref{fig:masking_example} b) and c), respectively.

% The magnitude filter $M_{||\cdot||}(\mathbf{x}_i)$ is calculated as the Heaviside function of the difference of the temporally averaged Euclidean norm  \(|\overrightarrow{v_i}|\) which quantifies the intensity of the motion of each voxel, and the threshold $T_{norm}$, which is defined as a selectable upper percentile (e.g., 40th, 50th, 60th, etc.) of the corresponding norm distribution:

% \begin{equation}
% \begin{aligned}
%     M_{||\cdot||}(\mathbf{x}_i) = H(|\overrightarrow{v}_{(\mathbf{x}_i)}| - T_{norm})  \\ 
%     \text{ with } T_{norm} \in [0,100]
% \end{aligned}    
% \end{equation}

% This ensures that only the voxel exhibiting strong motion are included, effectively excluding weak non-cardiac motion and noise while preserving the regions associated with the heart motion.

\subsection{Cardiac Keyframes}
\label{sec2_3:cardiac_key_frames}
% Maybe it makes more sense to put a major part of this to the Introduction
The cardiac cycle consists of alternating phases of contraction, referred to as systole, and relaxation, the diastole. Each of these phases is associated with specific mechanical events within the heart. Accurate identification of key time points in the cardiac cycle is of great importance for the evaluation of cardiac function. We are able to detect five time points, referred to as keyframes, from the motion descriptor, which are \acrshort{ed}, \acrshort{es}, \acrfull{ms}, \acrfull{pf}, and \acrfull{md}. 
%Furthermore, the  \acrshort{ed} and \acrshort{es} keyframes can be identified in accordance with the \acrshort{lv} blood pool volume curve, provided that segmentation is available. The only frame, which can be directly derived from the \acrshort{ecg} signal is the  \acrshort{ed} as it is associated with the R-wave peak, which is only feasible if the data is stored in conjunction with the \acrshort{cmr}. 
\acrshort{ed} is identified as the frame in the \acrshort{cmr} showing the largest ventricular volume, occurring just before the myocardium starts contracting. It can be derived from the \acrshort{lv} blood-pool volume curve as the global maximum.
The following \acrshort{ms} frame occurs during systole and represents the moment of maximum contractile motion. As the myocardium is actively contracting and pushing the most blood into the arteries, \acrshort{ms} is associated with the most pronounced reduction in volume. 
The \acrshort{es} is characterized by the maximum contraction of the myocardium and the closure of the semilunar valves. It is identified as the frame with the smallest ventricular volume, corresponding to the the global minimum of the \acrshort{lv} volume curve and occurring shortly after the T wave of the \acrshort{ecg}. 
The \acrshort{pf} occurs during early diastole when rapid ventricular filling takes place and corresponds to the frame with the strongest increase in \acrshort{lv} volume. The frame immediately preceding atrial contraction is identified as the \acrshort{md}, often observed as a decrease in atrial volume in \acrshort{4ch}  \acrshort{cmr} images or an additional \acrshort{lv} extension after slowing of filling in \acrshort{sax} \acrshort{cmr} images. Interestingly, we observed that we can identify these keyframes in the one-dimensional motion descriptor.

Based on a rule set, we leverage the 1D motion descriptor $\alpha_t$ to identify the five keyframes during the cardiac cycle. Given the variability of the initial cardiac phase in \acrshort{cmr} images  (Section \ref{sec24:dataset}), we first locate \acrshort{ms}, which corresponds to the global minimum of the contraction-relaxation curve. Subsequently, the remaining keyframes are determined by applying a sequence of rules to the cyclic sub-sequence (compare Figure \ref{fig:cardiaC_curves}). 

% including \acrshort{ed}, \acrfull{ms}, the maximum contractile motion causing peak ejection between \acrshort{ed} and \acrshort{es}, \acrshort{es}, \acrfull{pf}, which is the early relaxation of the diastole causing the maximum blood flow, and the \acrfull{md}, which is the time point right before atrial contraction at the onset of the p-wave. 

\begin{align*}
    & \textbf{\acrshort{ms}}=t_m                   \text{ where } \alpha(t_m) \leq \alpha(t)    & \text{ for } t \in T \\
    & \textbf{\acrshort{es}}=max\{\alpha(t)=0      \text{ and } \alpha' (t)>0\}                 & \text{ for } t \in [\acrshort{ms},\acrshort{pf}] \\
    & \textbf{\acrshort{pf}}=min\{\alpha'(t)=0     \text{ and } \alpha^{\prime\prime}(t)<0\}                & \text{ for } t \in [\acrshort{es},\acrshort{ms}] \\
    & \textbf{\acrshort{ed}}=max\{\alpha(t)=0      \text{ and } \alpha' (t)<0\}                 & \text{ for } t \in [\acrshort{pf},\acrshort{ms}] \\
    & \textbf{\acrshort{md}} = max\{\alpha'(t)=0   \text{ and } \alpha^{\prime\prime}(t)<0\}                & \text{ for } t \in [\acrshort{pf}, \acrshort{ed}] 
    % &\textbf{\acrshort{ms}}= { \acrshort{ed} + \acrshort{es}\over{ 2}}  \\
\end{align*}

\subsection{Datasets}
\label{sec24:dataset}
\begin{sidewaystable*}[]
    \centering
    \scriptsize
    \caption{Summary of datasets used, including patient groups (see Table \ref{tab:pathology}),  segmentation annotations, usage, imaging views (\acrshort{4ch}: four-chamber, \acrshort{sax}: short-axis), number of cases ("Num. cases") and keyframe annotations ("Keyframes") per view.  Segmentation annotations are published bi-ventricular segmentation at \acrshort{ed} and \acrshort{es}. "Keyframes" include either the original dataset annotations or additional physician annotations (asterisk*; Section \ref{sec2_3:cardiac_key_frames}). "All" denotes all five keyframes (\acrshort{ed}, \acrshort{ms}, \acrshort{es}, \acrshort{pf}, \acrshort{md}). Values in "Num. cases" and "Keyframes" match the order of listed views.}
    \label{tab:combined_dataset_pathologies}
    \begin{tabular}{p{2cm} p{4cm} p{2cm} p{2.3cm} p{1.5cm} p{1cm} p{1cm} p{1.5cm}}
        \hline
        Abbreviated dataset name & Full dataset name and citation  & Patient groups & Segmentation \newline(\acrshort{ed}, \acrshort{es})  & Usage & Views & Num. cases &   Keyframes  \\
        \hline
        \acrshort{mm2}\newline train & \acrlong{mm2} \citep{mnm2_3267857} & \acrshort{nor}, \acrshort{dlv}, \acrshort{hcm}, \acrshort{arr}, \acrshort{tof}, \acrshort{cia}, \acrshort{drv}, \acrshort{tri} & \acrshort{lv}, \acrshort{rv}, \acrshort{lv} MYO & Train/Test & \acrshort{4ch}; \acrshort{sax} & 200;\newline200  & \acrshort{ed}, \acrshort{es};\newline\acrshort{ed}, \acrshort{es}   \\
        \hline
        \acrshort{mm2}\newline test & \acrlong{mm2} \citep{mnm2_3267857} & \acrshort{nor}, \acrshort{dlv}, \acrshort{hcm}, \acrshort{arr}, \acrshort{tof}, \acrshort{cia}, \acrshort{drv}, \acrshort{tri}  & \acrshort{lv}, \acrshort{rv}, \acrshort{lv} MYO & Test & \acrshort{4ch}; \acrshort{sax} & 160;\newline160  & All*;\newline \acrshort{ed}, \acrshort{es} \\
        \hline
        \acrshort{mnms} & \acrlong{mnms} \citep{mnm2_9458279} & \acrshort{nor}, \acrshort{dcm}, \acrshort{hcm}, \acrshort{arr}, Others & \acrshort{lv}, \acrshort{rv}, \acrshort{lv} MYO & Test & \acrshort{sax} & 345 & \acrshort{ed}, \acrshort{es} \\
        \hline
        \acrshort{acdc} & \acrlong{acdc} \citep{acdc_dataset_8360453} & \acrshort{nor}, \acrshort{dcm}, \acrshort{hcm}, \acrshort{arr}, \acrshort{minf}  & \acrshort{lv}, \acrshort{rv}, \acrshort{lv} MYO  & Test & \acrshort{sax} & 100 & All* \\
        \hline
        \acrshort{gcn} & \acrlong{gcn} \citep{gcn_sarikouch2011}  & \acrshort{tof} & None & Test & \acrshort{4ch}; \acrshort{sax} & 206;\newline265 & \acrshort{ed}, \acrshort{es}*;\newline All* \\
        \hline
    \end{tabular}
\end{sidewaystable*}

\begin{table}[ht]
    \centering
\scriptsize
    \caption{Overview of pathologies. "Others" encompasses a range of less common or mixed cardiac conditions, including Hypertensive Heart Disease (HHD), Abnormal Right Ventricle (ARV), Athlete Heart Syndrome (AHS), Ischaemic Heart Disease (IHD), Left Ventricle Non-Compaction (LVNC), and other atypical or unclassified cardiomyopathies.}
    \label{tab:pathology}
    \begin{tabular}{l| c|c }
         Abbreviation & Pathology & n  \\
         \hline
         \acrshort{arr} & \acrlong{arr} &  30  \\
         \acrshort{arv} & \acrlong{arv} &  34  \\
         \acrshort{cia} & \acrlong{cia} &  30  \\
         \acrshort{dcm} & \acrlong{dcm} &  117  \\                  
         \acrshort{dlv} & \acrlong{dlv} &  55  \\
         \acrshort{drv} & \acrlong{drv} &  25  \\
         \acrshort{hcm} & \acrlong{hcm} &  160  \\
         \acrshort{minf} & \acrlong{minf} &  20  \\
         \acrshort{nor} & \acrlong{nor} &  179  \\
         \acrshort{tof} & \acrlong{tof} &  295  \\
         \acrshort{tri} & \acrlong{tri} &  25  \\
         Other & - & 59  \\
    \end{tabular}
\end{table}

We utilized 4 datasets for development and testing of our proposed method,  namely  \acrshort{mnms}  \citep{mnm2_9458279}, \acrshort{mm2} \citep{ mnm2_9458279, mnm2_3267857}, \acrshort{acdc} \citep{acdc_dataset_8360453}, which are publicly available, and \acrshort{gcn} \citep{gcn_sarikouch2011}. The datasets are summarized in Table \ref{tab:combined_dataset_pathologies} and described in more detail in the \ref{sec:App-dataset}. 
All dataset except \acrshort{gcn} contain bi-ventricular segmentation at \acrshort{ed} and \acrshort{es} and build on top of the \acrshort{acdc} challenge \acrfull{sop}. 

The training of the deformable registration model for keyframe detection, as well as the segmentation model for anatomical focus points, was conducted using the 200 cases from the \acrshort{mm2} training subset. The keyframe detection was performed on both the training subset and the remaining cases. Two physicians annotated all five keyframes of the \acrshort{sax} images of \acrshort{acdc} and \acrshort{gcn}, and of the \acrshort{4ch} images of the \acrshort{mm2} test subset to expand the number of annotated cardiac keyframes. They followed the definition of keyframes as described in chapter \ref{sec2_3:cardiac_key_frames}. The provision of these additional annotations enables the assessment of inter-observer variability for  \acrshort{ed} and \acrshort{es} frames in comparison to the published annotations. The additional keyframe labels will be released on our GitHub repository.

\subsection{Evaluation Metrics}

To evaluate keyframe detection, we use the previously introduced \acrfull{cfd} \citep{koehler2022self}, an extension of the average Frame difference ($aFD$) that accounts for the cyclic nature of a potential keyframe distribution. The \acrshort{cfd} measures the minimum temporal difference between a ground truth keyframe  $p_i$ and its corresponding prediction  $\hat{p}_i$, considering the cyclic boundary conditions.

\begin{equation}
    cFD(p_i,\hat{p}_i) = min(|p_i-\hat{p}_i|, T - max(p_i, \hat{p}_i) + min(p_i, \hat{p}_i)
\end{equation}

where,  $i \in [ \acrshort{ed}, \acrshort{ms}, \acrshort{es}, \acrshort{pf}, \acrshort{md}]$ denotes the keyframe type, and $T$ is the total number of frames in the \acrshort{cmr}. This formulation accounts for edge cases where a keyframe is annotated at the start or end of the cycle, while the prediction occurs at the corresponding opposite boundary of the sequence.
The \acrfull{iov} refers to the difference between the original ground truth label and the annotation provided by our physicians, as measured by the \acrshort{cfd}. 
% We define a threshold value as the mean plus two standard deviations (SD) of the average \acrshort{iov}. Cases with a \acrshort{cfd} greater than this value are considered outliers.

% A novel quantification metric for the motion descriptor curve was introduced to assess the quality of the extracted motion information. This metric, inspired by physiological patterns of a healthy heart, was evaluated on healthy subjects from the \acrshort{mm2} dataset for both \acrshort{sax} and \acrshort{4ch} motion descriptors. Notably, it can be employed as an indicator of motion quality even in the absence of ground truth annotations, potentially aiding in the detection of pathological patterns. The metric leverages phase deviation and consistency of the cardiac cycle, including systole, diastole, and peak blood flow. Lower metric values indicate better motion quality. 
% To further validate the proposed metric, a correlation analysis was conducted between frame detection accuracy and motion descriptor quality.

\subsection{Experimental Setup}
\label{chapter2_6:experimental_setup}
Each model was trained on the trainings subset of the \acrshort{mm2} dataset (Section \ref{sec24:dataset}). To achieve the unified temporal length for the input $I$, we repeated $I_t$ along $t$ until we reached the network's input size of 40. Furthermore we linear interpolate $I$ to the respective target input spacing of $2.5 \text{mm}^3$ and $1.0 \text{mm}^2$ for \acrshort{sax} and \acrshort{4ch} respectively.

We compare our proposed method with a supervised \acrshort{lv}-volume based approach on the same data and refer to it as \textit{base}. For this we train a segmentation model on the  \acrshort{mm2} training dataset for each view to establish a baseline comparison. The \acrshort{lv} blood pool label was used to derive the \acrshort{lv} volume curve, from which the  \acrshort{ed} and \acrshort{es} frames were identified as the frames corresponding to the minimum and maximum volume, respectively. This approach was extended to the \acrshort{4ch} view, recognising that it primarily represents an \acrshort{lv}-area curve. However, since the relative changes in the curve are more relevant than the absolute volume values, the \acrshort{lv} area curve was treated similarly to the volume curve for keyframe detection.

In our self-supervised approach, the focus point $C$ is defined as the centre of mass of the computed mask $M$, averaged along the temporal axis, denoted as $C_{mse}$.  Four experimental settings were conducted to assess the impact of different focus point and its sensitivity to variations in relation to keyframe detection.
For comparison with $C_{mse}$, one other self-supervised focus point $C_{vol}$ and two supervised anatomically derived focus points $C_{lv}$ and $C_{sept}$ were defined.
$C_{vol}$ is defined as the centre of the entire \acrshort{cmr}-volume/-image. %Furthermore, the segmentation model was utilized to establish the calculation of anatomical focus points in a semi-supervised manner. 
The anatomical focus points are derived from the predicted segmentations, where $C_{lv}$ is defined as the centre of mass of the \acrshort{lv} blood-pool and $C_{sept}$ as the mean septum landmark (midpoint between the average anterior and inferior right ventricular insertion points (RVIP) \citep{koehler2022comparison}).

To ensure effective masking of non-cardiac motion and noise, suitable thresholds for both the magnitude of motion $T_{norm}$ and temporal directional change $T_{\Delta\alpha}$ were empirically determined. The optimal combination for \acrshort{sax} images was found to be $T_{norm} = 50^{th}$ and $T_{\Delta\alpha} = 0.8$, and for \acrshort{4ch} sequences $T_{norm} = 50^{th}$ and $T_{\Delta\alpha} = 1.2$.

\section{Results}
\label{sec4:ExperimentResults}

\begin{sidewaystable*}[]
\centering
\scriptsize
\caption{Cyclic frame difference (mean ± SD) for the \acrshort{sax} view for five datasets (\acrshort{mnms}, \acrshort{mm2} test and training, \acrshort{acdc}, \acrshort{gcn}) with respect to different focus points $C_n$. $base$ - supervised volume-based approach, $C_{mse}$ and $C_{vol}$ are computed fully self-supervised without prior anatomical knowledge, while $C_{sept}$  and $C_{lv}$ uses the segmentation model to compute anatomical focus point. Best results are marked in bold. The \acrshort{iov} is calculated between public annotations and our annotations. The first row of the \acrshort{mnms} dataset include the results reported by \cite{garcia2023cardiac}. $ \ast: p < 0.05$,  $\ast\ast: p < 0.01$ (vs. \textit{base}, paired Wilcoxon test). NR - not reported.}
\label{tbl:sax_key_frame}
\begin{tabular}{cccccccc}
\hline
\textbf{Data}  & \textbf{C\_n} & \textbf{all} & \textbf{ \acrshort{ed}} & \textbf{\acrshort{ms}} & \textbf{\acrshort{es}} & \textbf{\acrshort{pf}} & \textbf{\acrshort{md}} \\ 
\hline\multirow{5}{*}{\acrshort{mnms}}
    & Garcia-Cabrera &  1.73  $\pm$  NR & 1.70   $\pm$  NR & - &  1.75  $\pm$  NR & - & - \\ 
    & $base$ &   1.84  $\pm$  2.21 & 1.76  $\pm$  2.17 & - & 1.92  $\pm$  2.25 & - & - \\
    & $C_{mse}$ & \textbf{1.28  $\pm$  1.60} & 1.01  $\pm$  1.36 $\ast\ast$ & - & \textbf{ 1.55  $\pm$  1.83} &-&-\\
    & $C_{vol}$ & 1.29  $\pm$  1.59 & 1.01  $\pm$  1.36 $\ast\ast$ & - & 1.56  $\pm$  1.82  & - & - \\
    & $C_{sept}$ & 1.35  $\pm$  1.65 & 0.96  $\pm$  1.22 $\ast\ast$ & - & 1.74  $\pm$  2.08  & - & - \\
    & $C_{lv}$ & 1.31  $\pm$  1.55 &  \textbf{0.94  $\pm$  1.10} $\ast\ast$ & - &  1.67  $\pm$  1.99 & - & - \\
\hline
\multirow{5}{*}{\shortstack{\acrshort{mm2} \\ train}}
    & $base$ & 1.56  $\pm$  1.59 & 1.47  $\pm$  1.54 & - & 1.66  $\pm$  1.64 & - & - \\
    & $C_{mse}$ &  \textbf{0.77 $\pm$  0.99 } & \textbf{0.67  $\pm$  1.07} $\ast\ast$ & - &  0.87  $\pm$   0.92 $\ast\ast$ &-&-\\
    & $C_{vol}$ & 0.81  $\pm$ 1.07 & 0.77 $\pm$  1.18 $\ast\ast$ & - & \textbf{0.86  $\pm$  0.97} $\ast\ast$ & - & - \\
    & $C_{sept}$ & 1.21 $\pm $ 1.48 & 1.22  $\pm$  1.67 $\ast$ & - & 1.21  $\pm$  1.28 $\ast\ast$ & - & - \\
    & $C_{lv}$ & 1.35 $\pm $ 1.61 & 1.49  $\pm$  1.92 & - & 1.21 $\pm$ 1.30 $\ast\ast$ & - & - \\
\hline
\multirow{5}{*}{\shortstack{\acrshort{mm2} \\ test}}  
    & $base$ & 1.68 $\pm$  2.18 & 1.75 $\pm$  2.46 & - &1.60$ \pm$ 1.90 & - & - \\
    & $C_{mse}$ & 1.05  $\pm$  1.41 & \textbf{ 0.96$ \pm$ 1.41} $\ast\ast$ & - & 1.14$ \pm$ 1.41 $\ast\ast$ & - & - \\
    & $C_{vol}$ & \textbf{1.01  $\pm$  1.32 }& 1.01$ \pm$ 1.33 $\ast\ast$ & - & \textbf{1.02$ \pm$ 1.31} $\ast\ast$ & - & - \\
    & $C_{sept}$ & 1.39  $\pm$  1.77 & 1.44$ \pm$ 1.86 & - &1.34 $\pm$  1.67 & - & - \\
    & $C_{lv}$ & 1.55  $\pm$  2.00 & 1.77$ \pm$ 2.35 & - &1.33 $\pm$  1.61 & - & - \\
\hline
\multirow{5}{*}{\acrshort{acdc}}  
    & $base$ & 1.81$ \pm$ 2.24 &1.55  $\pm$  2.12 & - & 2.08  $\pm$  2.36 & - & - \\
    & $C_{mse} $&\textbf{ 1.31$ \pm$  1.43 }&\textbf{0.94  $\pm$  1.32 } $\ast$ & 1.13  $\pm$  1.02 &\textbf{ 1.16$ \pm$ 1.12 } $\ast\ast$ &\textbf{ 1.82  $\pm$  2.13} &\textbf{ 1.49  $\pm$  1.59} \\
    & $C_{vol}$ & 1.64$ \pm$ 2.22 &1.27$ \pm$ 2.36 & 1.41  $\pm$  1.66 & 1.35$ \pm$ 1.67 $\ast$ & 2.19  $\pm$  2.61 & 1.98  $\pm$  2.78 \\
    & $C_{sept} $& 1.81$ \pm$ 2.00 & 1.77$ \pm$ 2.39 & 1.05  $\pm$  0.93 & 1.63 $\pm$ 2.13 $\ast$ & 2.31  $\pm$   2.37 & 2.23  $\pm$  2.62 \\
    & $C_{lv} $& 1.78$ \pm$ 2.00 &1.95$ \pm$ 2.47 &\textbf{ 1.12  $\pm$  0.93 }& 1.47$ \pm$ 1.89 $\ast$ &  2.04  $\pm$   1.97 & 2.33  $\pm$  2.72 \\
    \cmidrule{2-8}
    & \textit{\acrshort{iov}} & 0.99$ \pm$  1.23 & 1.07$ \pm$  0.86 & - & 0.91  $\pm$  1.60 & - & - \\
\hline
\multirow{5}{*}{\acrshort{gcn}}  
    & $base$ & $2.06 \pm  1.29$ & $1.35 \pm 1.41$ & -  & $2.78 \pm 1.18$ & - & - \\
    & $C_{mse} $& \textbf{1.00 $\pm $ 0.58} & $0.97 \pm 0.76$  $\ast$ & \textbf{0.87  $\pm$  0.57} & \textbf{0.98  $\pm$  0.39}  $\ast\ast$ & \textbf{1.18  $\pm$  0.54} & \textbf{1.02 $\pm$ 0.64} \\
    & $C_{vol}$ & $1.05 \pm 0.63$ & \textbf{0.93  $\pm$   0.75}  $\ast\ast$ & $1.03 \pm  0.70$ & $1.05 \pm 0.46$  $\ast\ast$ & $1.19 \pm 0.56$ & $1.03 \pm 0.67$ \\
    & $C_{sept} $& $1.44 \pm 1.00$ & $1.03 \pm 0.82$ $\ast$ & $1.28 \pm 0.94$ & $1.40 \pm  0.82$ $\ast\ast$  & $2.15 \pm 0.77$ & $1.30 \pm  1.24$ \\
    & $C_{lv}$ & $1.70 \pm 1.10$ & $1.05 \pm 1.27$  $\ast\ast$  & $1.89 \pm 1.07$ & $1.74 \pm  0.86$  $\ast\ast$ & $2.28 \pm 0.85$ & $1.53 \pm 1.45$ \\
\hline
\end{tabular}
\end{sidewaystable*}

\begin{sidewaystable*}[]
\centering
\scriptsize
\caption{Cyclic frame difference (mean ± SD) for the \acrshort{4ch} view for different datasets (\acrshort{mm2} test and training dataset and \acrshort{gcn}), with respect to different focus points $C_n$. $base$ - supervised volume-based approach, $C_{mse}$ and $C_{vol}$ are computed fully self-supervised without prior anatomical knowledge, while $C_{sept}$  and $C_{lv}$ uses the segmentation model to compute anatomical focus point. Best results are marked in bold. In case of the \acrshort{mm2} test dataset, the mean \acrshort{cfd} would be 0.82  $\pm$  0.89 if only  \acrshort{ed} and \acrshort{es} are considered like it is the case for base. The \acrfull{iov} is calculated between public annotations and our annotations. $ \ast: p < 0.05$,  $\ast\ast: p < 0.01$ (vs. \textit{base}, paired Wilcoxon test). } 
\begin{tabular}{cccccccc}
\hline
\textbf{Data}  & \textbf{C\_n} & \textbf{all} & \textbf{ \acrshort{ed}} & \textbf{\acrshort{ms}} & \textbf{\acrshort{es}} & \textbf{\acrshort{pf}} & \textbf{\acrshort{md}} \\ 
\hline
\multirow{5}{*}{\shortstack{\acrshort{mm2} \\ train}}  
& $base$ &   1.30  $\pm$  1.34 & 1.33  $\pm$  1.59 & - & 1.27  $\pm$  1.10  & - & - \\
    & $C_{mse}$ & 1.21  $\pm$  1.37 & 1.23  $\pm$  1.53  & - & 1.20  $\pm$  1.21 &  - & - \\
    & $C_{vol}$ &  1.93  $\pm$  2.45 &  2.04  $\pm$  2.58 & - &  1.81  $\pm$  2.32 & - & - \\
    & $C_{sept}$ & \textbf{1.17  $\pm$  1.36} & \textbf{1.18  $\pm$  1.48 } & - & 1.16  $\pm$  1.23  & - & - \\
    & $C_{lv}$ & 1.17  $\pm$  1.45   & 1.23  $\pm$  1.50 & - & \textbf{1.12  $\pm$  1.40} $\ast$ & - & - \\
\hline
\multirow{5}{*}{\shortstack{\acrshort{mm2} \\ test}}    
    & $base$ &   0.92  $\pm$  1.20 & 0.93  $\pm$  1.20 & -  & \textbf{0.91  $\pm$  1.20} & - & - \\
    & $C_{mse}$ &  1.27  $\pm$  1.32  & 0.94  $\pm$  1.11 & 1.28  $\pm$  1.04 & 0.99  $\pm$  1.09 & 1.51  $\pm$  1.61 & 1.61  $\pm$  1.75 \\
    & $C_{vol}$ &  1.81  $\pm$  2.17 & 1.51  $\pm$  2.05 & 1.70  $\pm$  2.02 & 1.59  $\pm$  2.19 & 2.13  $\pm$  2.37 & 2.09  $\pm$  2.23 \\
    & $C_{sept}$ &   \textbf{1.15  $\pm$  1.14} & \textbf{0.72  $\pm$  0.78} & 1.39  $\pm$  1.08 & 0.93  $\pm$  1.01 & \textbf{1.38  $\pm$  1.35 }& \textbf{1.34  $\pm$  1.50} \\
    & $C_{lv}$ & 1.17  $\pm$  1.22 &  0.84  $\pm$  0.93 & \textbf{1.21  $\pm$  1.01} & 0.96  $\pm$  1.11 & 1.41  $\pm$  1.54 & 1.44  $\pm$  1.49 \\
    \cmidrule{2-8}
    & \textit{\acrshort{iov}} & 1.17  $\pm$  1.58 & 1.13  $\pm$  1.64 & - & 1.20  $\pm$  1.51 & - & - \\ 
\hline
\multirow{5}{*}{\acrshort{gcn}}
    & $base$        & 3.26  $\pm$  3.19   & 3.40  $\pm$  3.44 & - & 3.12  $\pm$  2.93  &  - & - \\
    & $C_{mse} $    & 1.73  $\pm$  1.94   & 1.78  $\pm$  2.04  $\ast\ast$ & - & 1.67  $\pm$  1.83 $\ast\ast$  &  - & - \\
    & $C_{vol}$     & 2.10  $\pm$  2.31  & 2.45  $\pm$  2.35  $\ast$ & - & 1.75  $\pm$  2.27  $\ast\ast$ & - & - \\
    & $C_{sept} $   & \textbf{1.58  $\pm$  1.91 }  & \textbf{1.49  $\pm$  1.97}  $\ast\ast$ & - & 1.67  $\pm$  1.75  $\ast\ast$ &  - & - \\
    & $C_{lv}$      & 1.75  $\pm$  2.02   & 1.89  $\pm$  2.14  $\ast\ast$ & - &  \textbf{1.61  $\pm$  1.90} $\ast\ast$  &  - & - \\
\hline
\end{tabular}
\label{tbl:lax_key_frame}
\end{sidewaystable*}

\begin{figure}
    \centering
    \includegraphics[width=0.9\linewidth]{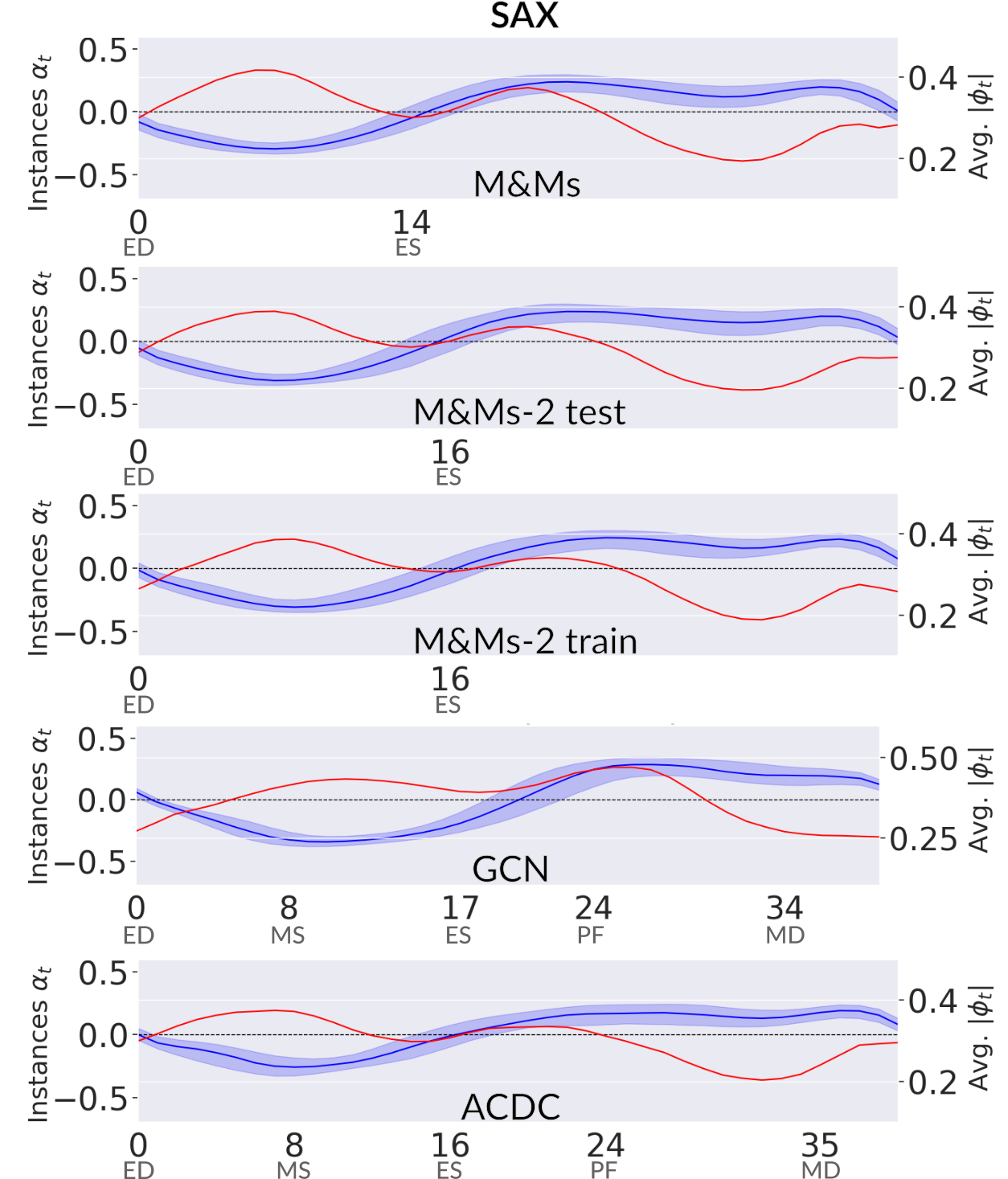}
    \caption{\textbf{Motion descriptor $\alpha_t$ for all five \acrshort{sax} datasets.} Each subplot shows the median of the masked $\alpha_t$ (blue/left axis)  with \acrshort{iqr} (light blue/left axis) of each dataset along with its median displacement magnitude $|\overrightarrow{v}_t|$ (red/right axis), which is normalized in a range of $[0,1]$.  Every input was linearly interpolated to 40 frames. The averaged phase indices (x-axis) are displayed together with the corresponding phase. In order to visualize the general properties the graph lines were aligned at the  \acrshort{ed} phase and resized, with the original data remained unaligned.}
    \label{fig:motioncurves_iqr_sax}
\end{figure}

\begin{figure}[ht!]
    \centering
    \includegraphics[width=0.9\linewidth]{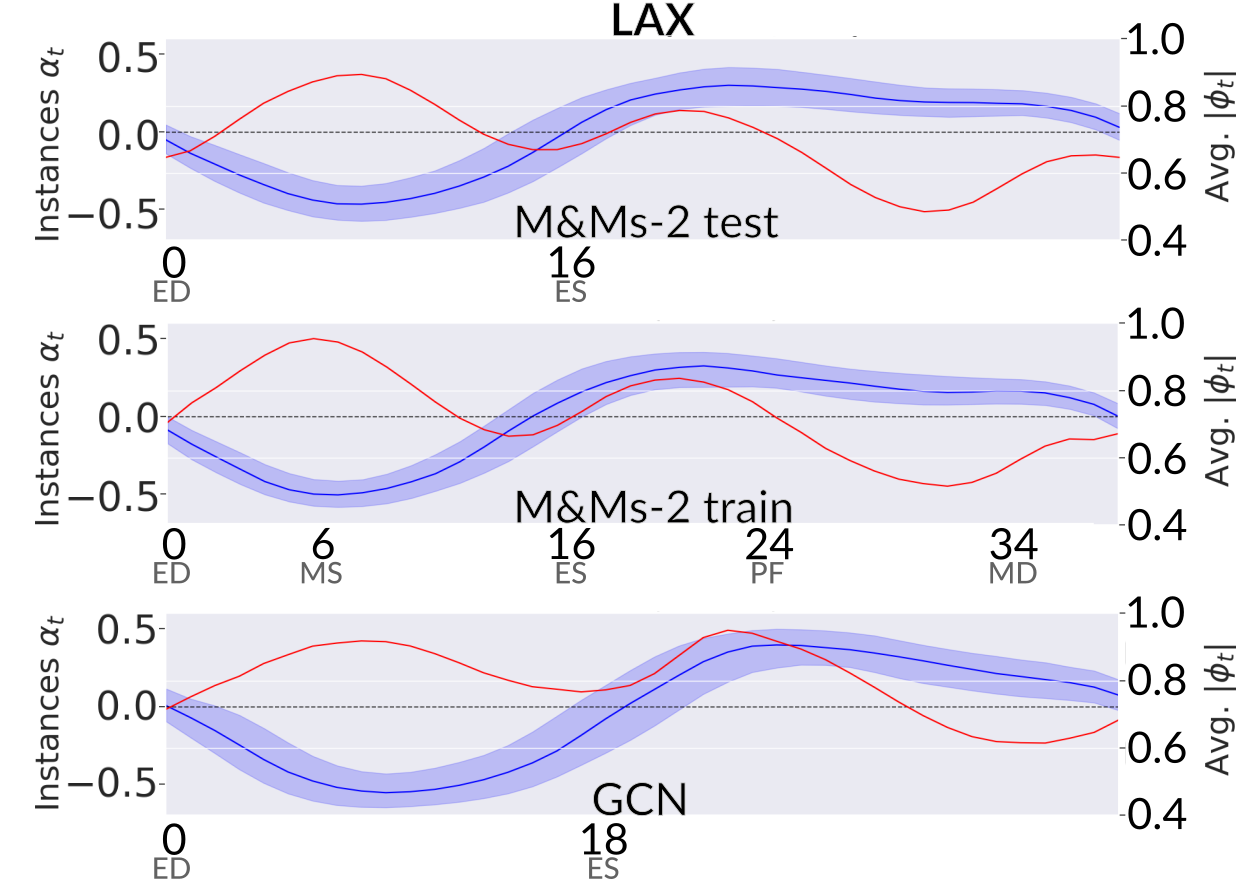}
    \caption{\textbf{Motion descriptor $\alpha_t$ for all three \acrshort{4ch} datasets [bottom].} Each subplot shows the median of the masked $\alpha_t$ (blue/left axis)  with \acrshort{iqr} (light blue/left axis) of each dataset along with its median displacement magnitude $|\overrightarrow{v}_t|$ (red/right axis), which is normalized in a range of $[0,1]$.  Every input was linearly interpolated to 40 frames. The averaged phase indices (x-axis) are displayed together with the corresponding phase. In order to visualize the general properties the graph lines were aligned at the  \acrshort{ed} phase and resized, with the original data remained unaligned.}
    \label{fig:motioncurves_iqr_las}
\end{figure}

Keyframe detection was performed for all dataset based on our novel motion descriptor, derived from dense deformable vector fields. The measured \acrshort{cfd} for each dataset and cardiac keyframe is presented in Table \ref{tbl:sax_key_frame} for \acrshort{sax} and Table \ref{tbl:lax_key_frame} for \acrshort{4ch}. To assess the sensitivity of \acrshort{cfd} to the choice of focus point $C_n$, multiple configurations are compared in the tables as described in Section \ref{chapter2_6:experimental_setup}. The performance from a baseline method deriving keyframes from the \acrshort{lv} volume curves is also reported in the tables, annotated as \textit{base}. The method uses the \acrshort{lv} volume curve to estimate \acrshort{ed} and \acrshort{es} from predicted segmentation masks as the maximum and minimum volume, respectively. We used the paired Wilcoxon test to calculate the statistical significance between the volume-base approach \textit{base} and our approach with different focus points, $C_n$. Notably, our approach outperforms the \textit{base} method in all cases except for \acrshort{es} in the \acrshort{mm2} test subset, where the difference was not significant ($p > 0.05$). When employing $C_{mse}$, the centre of mass of the mask from the self-supervised approach, our method achieved significantly improved keyframe detection for both ED and ES across all datasets in \acrshort{sax}. For \acrshort{lax}, significant improvements were observed only in the \acrshort{gcn} dataset.

\acrfull{iov} is listed in the respective \textit{\acrshort{iov}} row of the tables \ref{tbl:sax_key_frame} and \ref{tbl:lax_key_frame}. 
For \acrshort{4ch} cine \acrshort{cmr}, the combined mean \acrshort{cfd} for \acrshort{ed} and \acrshort{es} is  $1.16 \pm 1.45$, with maximum differences  of 12  and 6 frames in  \acrshort{ed} and \acrshort{es}, respectively. \acrshort{iov} is slightly lower in \acrshort{sax} cine \acrshort{cmr}, with a \acrshort{cfd} of $0.99 \pm 1.23$ and maximum frame differences of 6 and 10 frames for \acrshort{ed} and \acrshort{es}, respectively.  

The segmentation model achieved a DICE score of $0.90 \pm 0.07$ for \acrshort{lv} segmentation in \acrshort{sax} and of $0.90 \pm 0.10$ for \acrshort{4ch} in the 160 cases of the \acrshort{mm2} test dataset. The results of the segmentation model for all labels in both the training and test set are detailed in Table \ref{tbl:segmentation_dice}. 

The median motion descriptors with \acrfull{iqr} are illustrated in Figure \ref{fig:motioncurves_iqr_sax} and \ref{fig:motioncurves_iqr_las} for \acrshort{sax} at the top and for \acrshort{4ch} at the bottom for each cohort, including the median norm. 

\begin{figure}[ht]
    \centering
    \includegraphics[width=0.7\linewidth]{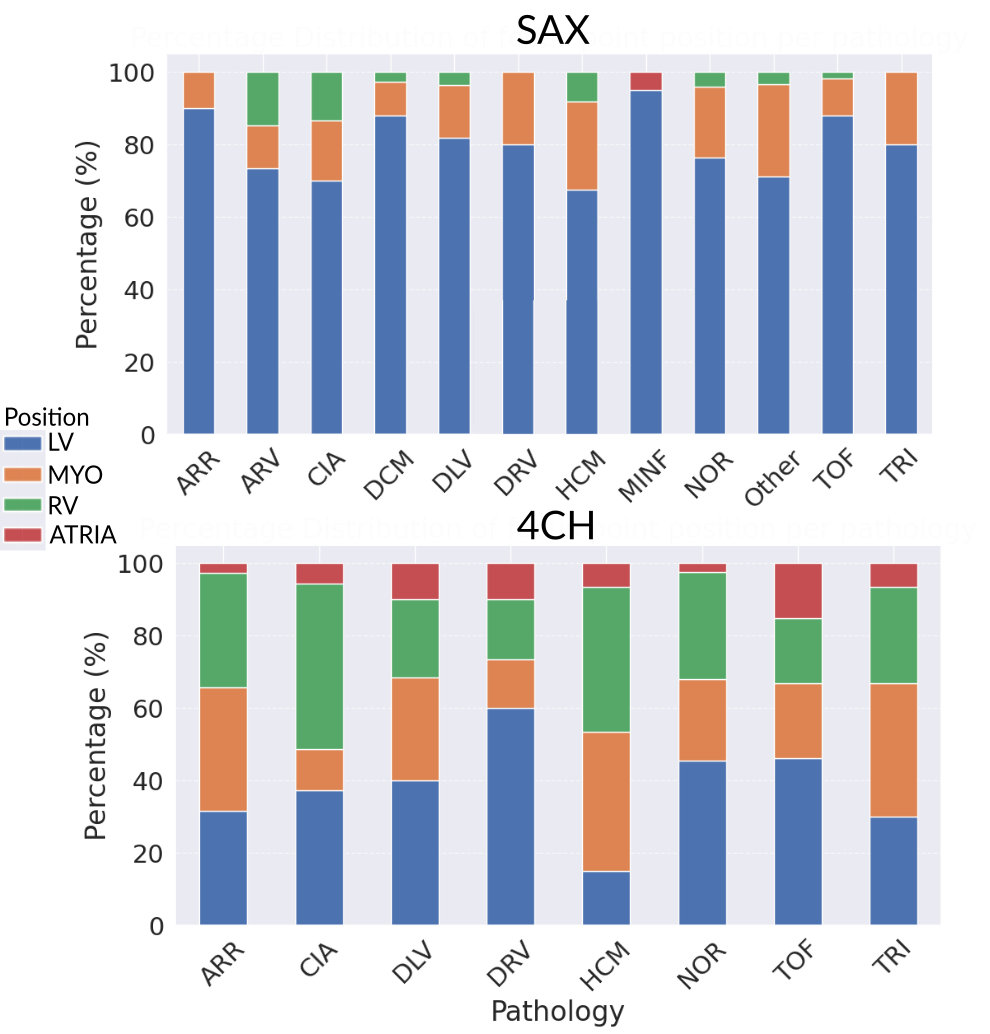}
    \caption{\textbf{Distribution of focus point $C_{mse}$ location in relation to bi-ventricular segmentation at \acrshort{ed} summarized across all dataset.} For a better overview, the position per pathology is shown as a percentage per anatomical structure: \acrshort{lv} (blue) - left ventricle, MYO (orange) - \acrshort{lv} myocardium, \acrshort{rv} (green) - right ventricle, ATRIA (red) - atria in \acrshort{4ch} and other in \acrshort{sax}.
    The cases which where not covered by the bi-ventricular segmentation mask were controlled manually. All $C_{mse}$ in  \acrshort{4ch} cases, which were outside the bi-ventricular segmentation mask, were found to be in location of the atria. For only one patient in the \acrshort{acdc} dataset with \acrshort{minf} the focus point $C_{mse}$ was computed directly outside the \acrshort{lv} myocardium.}
    \label{fig:focuspoint_location_distribution}
\end{figure}

The distribution of the location of the self-supervised computed focus point is illustrated in Figure \ref{fig:focuspoint_location_distribution}. For most of the \acrshort{sax} cases, the focus point is located inside the \acrshort{lv}. Only for one case $C_{mse}$ is located directly outside of the heart, near the \acrshort{lv} myocardium wall. For \acrshort{4ch} the location is more equally distributed, with some cases being located in the region of the atria.

\section{Discussion}
\label{sec5:discussion}

In this study, we have devised a methodology for the computation of a motion descriptor to express cardiac dynamics over time. It is based on the mean direction and norm of a sequential deformable registration field $\phi_t$ computed in a self-supervised manner in relation to varying focus points $C_n$.
A set of rules is defined, extending the state-of-the-art by extracting not only two but five cardiovascular keyframes in cine \acrshort{cmr} sequences with different views and of varying lengths independent of the starting phase. These rules are based on physiological principles and have been further optimized to achieve optimal performance for healthy hearts. They minimise the range of outliers while achieving consistent results in pathological cases. This approach prioritizes generalisability while maintaining high accuracy, even in the presence of potential cut-off sequences or pathological conditions. 

The results demonstrate that the proposed method generally outperforms segmentation-based detection, particularly in cases involving data from unseen scanners and clinics as in the \acrshort{acdc} and \acrshort{gcn} dataset. It is furthermore superior to segmentation-based approaches, as these models require manual ground truth labels to be trained on. 

The motion descriptor $\alpha$ (Figure \ref{fig:motioncurves_iqr_sax} and \ref{fig:motioncurves_iqr_sax}) displays a high degree of consistency with the typical cardiac characteristics with analogous patterns in both views, as hypothesised. The observed pattern in the curve, in which one third of the curve shows consistently negative values (indicating contractile motion) and the remaining two thirds positive values (indicating relaxing motion), reflects the typical cardiac cycle with systole and diastole. Therefore, the zero crossings indicate the end of each phase, consistent with the mean self-supervised detected \acrshort{ed} (<1.01 for \acrshort{sax}; <1.78 for \acrshort{4ch}) and \acrshort{es} (<1.55 for \acrshort{sax}; < 1.67 for \acrshort{4ch}). Compared to the volume-based method (\textit{base}), our approach yield significant improvements in \acrshort{sax} and results within the range of the \acrshort{iov}  or significantly improved in \acrshort{4ch}.

The global minimum, defined by the strongest motion direction towards the focus point, correlates with the \acrshort{ms} keyframe, where the contractile motion leads to the most pronounced reduction in volume. The detection of \acrshort{ms} is consistent across both views and datasets, with a mean \acrshort{cfd} below 1.22 frames, well within the range of typical \acrshort{iov}, underscoring the robustness of our approach.
The two maxima of the motion descriptor, defined as the points at which the majority of the voxels move away from the focus point, are indicative of the most potent relaxing motions. The more pronounced peak, which occurs shortly after the zero crossing (\acrshort{es}), has been shown to correlate with the peak flow. This is the point in time at which the heart undergoes its most significant expansion, which occurs immediately after the opening of the mitral and tricuspid valves. The less pronounced maximum is the point in time immediately following the contraction of the atria, which leads to a further slight expansion of the myocardium.  In contrast to the other keyframes, the identification of intermediate diastolic phases such as \acrshort{pf} and \acrshort{md} remains more challenging, particularly in pathological cases where relaxation patterns may be irregular, exhibiting either multiple diffuse peaks or a single dominant one. These complexities are reflected in the results, with mean \acrshort{cfd} for \acrshort{pf} ranging from 1.18 to 1.82 frames and for \acrshort{md} from 1.02 to 1.49, indicating a higher variability but still demonstrating competitive performance in the face of increased physiological ambiguity.

The observation that the second peak is less pronounced at LAX than at SAX can be attributed to the image section, which incorporates the atria at LAX, whereas this section is absent at SAX. As the atria contract, this contrary motion creates a negative direction that is the opposite of the positive direction value of the ventricles as they expand. In general, the slightly poorer results in \acrshort{4ch} images is generally attributable to the opposing motion phases of the atria and ventricles. As demonstrated in the motion field of Figure \ref{fig:masking_example}b, the atrial region exhibits contractile motion concurrently with the relaxation phase of the ventricles. This overlap may subtly affect the motion descriptor curve, thereby leading to a reduction in detection accuracy. This is also displayed in a broader \acrshort{iqr} of the motion descriptor as shown in Figure \ref{fig:motioncurves_iqr_las} in comparison to the slimmer \acrshort{iqr} of the \acrshort{sax} curves in Figure \ref{fig:motioncurves_iqr_sax}.

For \acrshort{sax}, our self-supervised approach significantly outperforms the supervised baseline across multiple datasets. Notably, on the \acrshort{mm2}$_{test}$ dataset, differences between the baseline and $C_{mse}$ are significant ($p < 0.01$, 1.05 $\pm$ 1.41 ($C_{mse}$) vs. 1.68 $\pm$ 2.18 ($base$)), and even more pronounced for \acrshort{mm2}$_{train}$ ($p < 0.0001$, 0.77 $\pm$ 0.99 ($C_{mse}$) vs. 1.56 $\pm$ 1.59 ($base$)) for both  \acrshort{ed} and \acrshort{es}. 

On the \acrshort{mnms} dataset, our method performs significantly better for \acrshort{ed}  ($p < 0.1e^{-3}$, 1.01 $\pm$ 1.36 ($C_{mse}$) vs. 1.76 $\pm$ 2.17 ($base$)), though the improvement for \acrshort{es} is marginal ($p=0.05$). Furthermore, our results surpass those of \cite{garcia2023cardiac}, who reported a \acrshort{afd} of 1.70 for  \acrshort{ed} and 1.75 for \acrshort{es}.

On the \acrshort{acdc} and \acrshort{gcn} dataset, our method shows significant improvements over the baseline for  \acrshort{ed} ($p < 0.05$, \acrshort{acdc}: 0.94 $\pm$ 1.32 ($C_{mse}$) vs. 1.55 $\pm$ 2.12 ($base$); \acrshort{gcn}: 1.00 $\pm$ 0.58 ($C_{mse}$) vs. 1.35 $\pm$ 1.41 ($base$)) and an even greater difference for \acrshort{es} ($p < 0.01$, \acrshort{acdc}: 1.16 $\pm$ 1.12 ($C_{mse}$) vs. 2.08 $\pm$ 2.36 ($base$); \acrshort{gcn}: 0.98 $\pm$ 0.39 ($C_{mse}$) vs. 2.78 $\pm$ 1.18 ($base$)). 

For \acrshort{4ch}, the improvements are more modest. On the \acrshort{mm2} datasets, our method slightly outperforms the baseline (\acrshort{mm2}$_{train}$: 1.21 $\pm$ 1.50 ($C_{lv}$) vs. 1.30 $\pm$ 1.34 ($base$); \acrshort{mm2}$_{test}$: 0.86 $\pm$ 0.97 ($C_{sept}$) vs. 0.92 $\pm$ 1.20 ($base$)), but the difference is not statistically significant. However, on the \acrshort{gcn} dataset the performance improves substantially ($p < 0.0001$, 1.58 $\pm$ 1.91 ($C_{sept}$) vs. 3.26 $\pm$ 3.19 ($base$)). In  cases where the self-supervised focus point $C_{mse}$ is outperformed by the anatomical focus points, the differences remain statistically non-significant($p > 0.05$). 

%The issue could be addressed by applying anatomically guided filtering to the motion fields, rather than relying solely on self-supervised methods. Furthermore, the incorporation of anatomical segmentation would facilitate the development of anatomically mapped motion descriptors. This could enhance the clinical relevance of the cardiac motion descriptor, offering valuable insights into functional abnormalities. This approach holds considerable promise for future research, with the potential to identify novel phenotypes and enhance diagnostic and prognostic capabilities.

Our approach enables accurate keyframe detection, including \acrshort{ed} and \acrshort{es}, as well as additional, less commonly analysed time points within the cardiac cycle. This temporal alignment allows for consistent and reproducible computation across patients and well defined phases, enhancing the interpretability and clinical relevance of the resulting motion descriptors. As demonstrated by \cite{koehler2025strain}, this phase-standardization supports meaningful inter-patient comparisons and much better discrimination between cohorts when performing aligned strain analysis - an approach that extends the current concept of strain considering further keyframes. %The alignment allowed essential meaningful assessment of cardiac function, reinforcing the relevance and strength of our approach in clinical and research settings.

\section{Conclusion}
\label{sec6:Conclusion}
We have introduced a fully self-supervised framework for detecting five cardiac keyframes in \acrshort{sax} and  \acrshort{4ch} \acrshort{cmr} cine sequences. Our framework has shown promising results that could allow its use in the clinical setting and save time in the diagnostic workflow.  In \acrshort{sax}, the average detection accuracy across all datasets  was within one frame for  \acrshort{ed} and under 1.16 frames  for \acrshort{es}. While the performance in  \acrshort{4ch} view was slightly lower, it remained within 1.50 frames for  \acrshort{ed} and 1.61 frames for \acrshort{es},  with the best results on the \acrshort{mm2} test dataset (0.87 for  \acrshort{ed} and 0.84 for \acrshort{es}).

Future work could be directed towards a more profound examination of the motion vector. This could include analysis of individual chamber movements after anatomical mapping of the motion vector $\alpha$. Furthermore, more \acrshort{lax} views could be incorporated, such as two- and three-chamber views. Overall, the approach could offer valuable insights into mechanical abnormalities at aligned phases of the cardiac cycle, with the potential to contribute towards identification of novel disease phenotypes.

\section*{Acknowledgements}
This study was funded by the Carl-Zeiss-Stiftung as part of the Multi-dimensionAI project (CZS-Project number: P2022-08-010) and data setes were received by the National Register for Congenital Heart Defects (Federal Ministry of Education and Research/grant number 01KX2140).

\newpage
\bibliographystyle{elsarticle-harv} 
\bibliography{main}

\begin{thebibliography}{31}
\expandafter\ifx\csname natexlab\endcsname\relax\def\natexlab#1{#1}\fi
\providecommand{\url}[1]{\texttt{#1}}
\providecommand{\href}[2]{#2}
\providecommand{\path}[1]{#1}
\providecommand{\DOIprefix}{doi:}
\providecommand{\ArXivprefix}{arXiv:}
\providecommand{\URLprefix}{URL: }
\providecommand{\Pubmedprefix}{pmid:}
\providecommand{\doi}[1]{\href{http://dx.doi.org/#1}{\path{#1}}}
\providecommand{\Pubmed}[1]{\href{pmid:#1}{\path{#1}}}
\providecommand{\bibinfo}[2]{#2}
\ifx\xfnm\relax \def\xfnm[#1]{\unskip,\space#1}\fi
%Type = Inproceedings
\bibitem[{Balakrishnan et~al.(2018)Balakrishnan, Zhao, Sabuncu, Guttag and Dalca}]{Balakrishnan_2018_CVPR}
\bibinfo{author}{Balakrishnan, G.}, \bibinfo{author}{Zhao, A.}, \bibinfo{author}{Sabuncu, M.R.}, \bibinfo{author}{Guttag, J.}, \bibinfo{author}{Dalca, A.V.}, \bibinfo{year}{2018}.
\newblock \bibinfo{title}{An unsupervised learning model for deformable medical image registration}, in: \bibinfo{booktitle}{Proceedings of the IEEE Conference on Computer Vision and Pattern Recognition (CVPR)}, pp. \bibinfo{pages}{9252--9260}.
%Type = Article
\bibitem[{Barcaro et~al.(2008)Barcaro, Moroni and Salvetti}]{barcaro_echo}
\bibinfo{author}{Barcaro, U.}, \bibinfo{author}{Moroni, D.}, \bibinfo{author}{Salvetti, O.}, \bibinfo{year}{2008}.
\newblock \bibinfo{title}{Automatic computation of left ventricle ejection fraction from dynamic ultrasound images}.
\newblock \bibinfo{journal}{Pattern Recognition and Image Analysis} \bibinfo{volume}{18}, \bibinfo{pages}{351--358}.
%Type = Article
\bibitem[{Bernard et~al.(2018)Bernard, Lalande, Zotti, Cervenansky, Yang, Heng, Cetin, Lekadir, Camara, Gonzalez~Ballester, Sanroma, Napel, Petersen, Tziritas, Grinias, Khened, Kollerathu, Krishnamurthi, Rohé, Pennec, Sermesant, Isensee, Jäger, Maier-Hein, Full, Wolf, Engelhardt, Baumgartner, Koch, Wolterink, Išgum, Jang, Hong, Patravali, Jain, Humbert and Jodoin}]{acdc_dataset_8360453}
\bibinfo{author}{Bernard, O.}, \bibinfo{author}{Lalande, A.}, \bibinfo{author}{Zotti, C.}, \bibinfo{author}{Cervenansky, F.}, \bibinfo{author}{Yang, X.}, \bibinfo{author}{Heng, P.A.}, \bibinfo{author}{Cetin, I.}, \bibinfo{author}{Lekadir, K.}, \bibinfo{author}{Camara, O.}, \bibinfo{author}{Gonzalez~Ballester, M.A.}, \bibinfo{author}{Sanroma, G.}, \bibinfo{author}{Napel, S.}, \bibinfo{author}{Petersen, S.}, \bibinfo{author}{Tziritas, G.}, \bibinfo{author}{Grinias, E.}, \bibinfo{author}{Khened, M.}, \bibinfo{author}{Kollerathu, V.A.}, \bibinfo{author}{Krishnamurthi, G.}, \bibinfo{author}{Rohé, M.M.}, \bibinfo{author}{Pennec, X.}, \bibinfo{author}{Sermesant, M.}, \bibinfo{author}{Isensee, F.}, \bibinfo{author}{Jäger, P.}, \bibinfo{author}{Maier-Hein, K.H.}, \bibinfo{author}{Full, P.M.}, \bibinfo{author}{Wolf, I.}, \bibinfo{author}{Engelhardt, S.}, \bibinfo{author}{Baumgartner, C.F.}, \bibinfo{author}{Koch, L.M.}, \bibinfo{author}{Wolterink, J.M.}, \bibinfo{author}{Išgum, I.}, \bibinfo{author}{Jang, Y.},
  \bibinfo{author}{Hong, Y.}, \bibinfo{author}{Patravali, J.}, \bibinfo{author}{Jain, S.}, \bibinfo{author}{Humbert, O.}, \bibinfo{author}{Jodoin, P.M.}, \bibinfo{year}{2018}.
\newblock \bibinfo{title}{Deep learning techniques for automatic mri cardiac multi-structures segmentation and diagnosis: Is the problem solved?}
\newblock \bibinfo{journal}{IEEE Transactions on Medical Imaging} \bibinfo{volume}{37}, \bibinfo{pages}{2514--2525}.
\newblock \DOIprefix\doi{10.1109/TMI.2018.2837502}.
%Type = Article
\bibitem[{Campello et~al.(2021)Campello, Gkontra, Izquierdo, Martín-Isla, Sojoudi, Full, Maier-Hein, Zhang, He, Ma, Parreño, Albiol, Kong, Shadden, Acero, Sundaresan, Saber, Elattar, Li, Menze, Khader, Haarburger, Scannell, Veta, Carscadden, Punithakumar, Liu, Tsaftaris, Huang, Yang, Li, Zhuang, Viladés, Descalzo, Guala, Mura, Friedrich, Garg, Lebel, Henriques, Karakas, Çavuş, Petersen, Escalera, Seguí, Rodríguez-Palomares and Lekadir}]{mnm2_9458279}
\bibinfo{author}{Campello, V.M.}, \bibinfo{author}{Gkontra, P.}, \bibinfo{author}{Izquierdo, C.}, \bibinfo{author}{Martín-Isla, C.}, \bibinfo{author}{Sojoudi, A.}, \bibinfo{author}{Full, P.M.}, \bibinfo{author}{Maier-Hein, K.}, \bibinfo{author}{Zhang, Y.}, \bibinfo{author}{He, Z.}, \bibinfo{author}{Ma, J.}, \bibinfo{author}{Parreño, M.}, \bibinfo{author}{Albiol, A.}, \bibinfo{author}{Kong, F.}, \bibinfo{author}{Shadden, S.C.}, \bibinfo{author}{Acero, J.C.}, \bibinfo{author}{Sundaresan, V.}, \bibinfo{author}{Saber, M.}, \bibinfo{author}{Elattar, M.}, \bibinfo{author}{Li, H.}, \bibinfo{author}{Menze, B.}, \bibinfo{author}{Khader, F.}, \bibinfo{author}{Haarburger, C.}, \bibinfo{author}{Scannell, C.M.}, \bibinfo{author}{Veta, M.}, \bibinfo{author}{Carscadden, A.}, \bibinfo{author}{Punithakumar, K.}, \bibinfo{author}{Liu, X.}, \bibinfo{author}{Tsaftaris, S.A.}, \bibinfo{author}{Huang, X.}, \bibinfo{author}{Yang, X.}, \bibinfo{author}{Li, L.}, \bibinfo{author}{Zhuang, X.}, \bibinfo{author}{Viladés, D.},
  \bibinfo{author}{Descalzo, M.L.}, \bibinfo{author}{Guala, A.}, \bibinfo{author}{Mura, L.L.}, \bibinfo{author}{Friedrich, M.G.}, \bibinfo{author}{Garg, R.}, \bibinfo{author}{Lebel, J.}, \bibinfo{author}{Henriques, F.}, \bibinfo{author}{Karakas, M.}, \bibinfo{author}{Çavuş, E.}, \bibinfo{author}{Petersen, S.E.}, \bibinfo{author}{Escalera, S.}, \bibinfo{author}{Seguí, S.}, \bibinfo{author}{Rodríguez-Palomares, J.F.}, \bibinfo{author}{Lekadir, K.}, \bibinfo{year}{2021}.
\newblock \bibinfo{title}{Multi-centre, multi-vendor and multi-disease cardiac segmentation: The m\&ms challenge}.
\newblock \bibinfo{journal}{IEEE Transactions on Medical Imaging} \bibinfo{volume}{40}, \bibinfo{pages}{3543--3554}.
\newblock \DOIprefix\doi{10.1109/TMI.2021.3090082}.
%Type = Article
\bibitem[{Dalca et~al.(2019)Dalca, Balakrishnan, Guttag and Sabuncu}]{Dalca_2019}
\bibinfo{author}{Dalca, A.V.}, \bibinfo{author}{Balakrishnan, G.}, \bibinfo{author}{Guttag, J.}, \bibinfo{author}{Sabuncu, M.R.}, \bibinfo{year}{2019}.
\newblock \bibinfo{title}{Unsupervised learning of probabilistic diffeomorphic registration for images and surfaces}.
\newblock \bibinfo{journal}{Medical Image Analysis} \bibinfo{volume}{57}, \bibinfo{pages}{226–236}.
\newblock \URLprefix \url{http://dx.doi.org/10.1016/j.media.2019.07.006}, \DOIprefix\doi{10.1016/j.media.2019.07.006}.
%Type = Article
\bibitem[{Darvishi et~al.(2013)Darvishi, Behnam, Pouladian and Samiei}]{darvishi_echo}
\bibinfo{author}{Darvishi, S.}, \bibinfo{author}{Behnam, H.}, \bibinfo{author}{Pouladian, M.}, \bibinfo{author}{Samiei, N.}, \bibinfo{year}{2013}.
\newblock \bibinfo{title}{Measuring left ventricular volumes in two-dimensional echocardiography image sequence using level-set method for automatic detection of end-diastole and end-systole frames}.
\newblock \bibinfo{journal}{Research in Cardiovascular Medicine} \bibinfo{volume}{2}, \bibinfo{pages}{39--45}.
%Type = Article
\bibitem[{Dezaki et~al.(2018)Dezaki, Liao, Luong, Girgis, Dhungel, Abdi, Behnami, Gin, Rohling, Abolmaesumi et~al.}]{dezaki2018cardiac}
\bibinfo{author}{Dezaki, F.T.}, \bibinfo{author}{Liao, Z.}, \bibinfo{author}{Luong, C.}, \bibinfo{author}{Girgis, H.}, \bibinfo{author}{Dhungel, N.}, \bibinfo{author}{Abdi, A.H.}, \bibinfo{author}{Behnami, D.}, \bibinfo{author}{Gin, K.}, \bibinfo{author}{Rohling, R.}, \bibinfo{author}{Abolmaesumi, P.}, et~al., \bibinfo{year}{2018}.
\newblock \bibinfo{title}{Cardiac phase detection in echocardiograms with densely gated recurrent neural networks and global extrema loss}.
\newblock \bibinfo{journal}{IEEE transactions on medical imaging} \bibinfo{volume}{38}, \bibinfo{pages}{1821--1832}.
%Type = Inproceedings
\bibitem[{Fiorito et~al.(2018)Fiorito, {\O}stvik, Smistad, Leclerc, Bernard and Lovstakken}]{fiorito2018detection}
\bibinfo{author}{Fiorito, A.M.}, \bibinfo{author}{{\O}stvik, A.}, \bibinfo{author}{Smistad, E.}, \bibinfo{author}{Leclerc, S.}, \bibinfo{author}{Bernard, O.}, \bibinfo{author}{Lovstakken, L.}, \bibinfo{year}{2018}.
\newblock \bibinfo{title}{Detection of cardiac events in echocardiography using 3d convolutional recurrent neural networks}, in: \bibinfo{booktitle}{2018 IEEE International Ultrasonics Symposium (IUS)}, \bibinfo{organization}{IEEE}. pp. \bibinfo{pages}{1--4}.
%Type = Inproceedings
\bibitem[{Garcia-Cabrera et~al.(2023)Garcia-Cabrera, Curran, O'Connor and McGuinness}]{garcia2023cardiac}
\bibinfo{author}{Garcia-Cabrera, C.}, \bibinfo{author}{Curran, K.M.}, \bibinfo{author}{O'Connor, N.E.}, \bibinfo{author}{McGuinness, K.}, \bibinfo{year}{2023}.
\newblock \bibinfo{title}{Cardiac magnetic resonance phase detection using neural networks}, in: \bibinfo{booktitle}{2023 31st Irish Conference on Artificial Intelligence and Cognitive Science (AICS)}, \bibinfo{organization}{IEEE}. pp. \bibinfo{pages}{1--4}.
%Type = Article
\bibitem[{Gifani et~al.(2010)Gifani, Behnam, Shalbaf and Sani}]{gifani2010automatic}
\bibinfo{author}{Gifani, P.}, \bibinfo{author}{Behnam, H.}, \bibinfo{author}{Shalbaf, A.}, \bibinfo{author}{Sani, Z.}, \bibinfo{year}{2010}.
\newblock \bibinfo{title}{Automatic detection of end-diastole and end-systole from echocardiography images using manifold learning}.
\newblock \bibinfo{journal}{Physiological Measurement} \bibinfo{volume}{31}, \bibinfo{pages}{1091}.
%Type = Inproceedings
\bibitem[{Jaderberg et~al.(2015)Jaderberg, Simonyan, Zisserman and Kavukcuoglu}]{Jadeberger_2015_33ceb07b}
\bibinfo{author}{Jaderberg, M.}, \bibinfo{author}{Simonyan, K.}, \bibinfo{author}{Zisserman, A.}, \bibinfo{author}{Kavukcuoglu, K.}, \bibinfo{year}{2015}.
\newblock \bibinfo{title}{Spatial transformer networks}, in: \bibinfo{editor}{Cortes, C.}, \bibinfo{editor}{Lawrence, N.}, \bibinfo{editor}{Lee, D.}, \bibinfo{editor}{Sugiyama, M.}, \bibinfo{editor}{Garnett, R.} (Eds.), \bibinfo{booktitle}{Advances in Neural Information Processing Systems}, \bibinfo{publisher}{Curran Associates, Inc.}. pp. \bibinfo{pages}{2017--2025}.
\newblock \URLprefix \url{https://proceedings.neurips.cc/paper_files/paper/2015/file/33ceb07bf4eeb3da587e268d663aba1a-Paper.pdf}, \DOIprefix\doi{10.48550/arXiv.1506.02025}.
%Type = Inproceedings
\bibitem[{Kachenoura et~al.(2006)Kachenoura, Delouche, Herment, Frouin and Diebold}]{kachenoura2007automatic}
\bibinfo{author}{Kachenoura, N.}, \bibinfo{author}{Delouche, A.}, \bibinfo{author}{Herment, A.}, \bibinfo{author}{Frouin, F.}, \bibinfo{author}{Diebold, B.}, \bibinfo{year}{2006}.
\newblock \bibinfo{title}{Automatic detection of end systole within a sequence of left ventricular echocardiographic images using autocorrelation and mitral valve motion detection}, in: \bibinfo{booktitle}{2007 29th Annual International Conference of the IEEE Engineering in Medicine and Biology Society}, \bibinfo{organization}{IEEE}. pp. \bibinfo{pages}{4504--4507}.
%Type = Inproceedings
\bibitem[{Koehler et~al.(2022a)Koehler, Hussain, Hussain, Young, Sarikouch, Pickardt, Greil and Engelhardt}]{koehler2022self}
\bibinfo{author}{Koehler, S.}, \bibinfo{author}{Hussain, T.}, \bibinfo{author}{Hussain, H.}, \bibinfo{author}{Young, D.}, \bibinfo{author}{Sarikouch, S.}, \bibinfo{author}{Pickardt, T.}, \bibinfo{author}{Greil, G.}, \bibinfo{author}{Engelhardt, S.}, \bibinfo{year}{2022}a.
\newblock \bibinfo{title}{Self-supervised motion descriptor for cardiac phase detection in 4d cmr based on discrete vector field estimations}, in: \bibinfo{booktitle}{International Workshop on Statistical Atlases and Computational Models of the Heart}, \bibinfo{organization}{Springer}. pp. \bibinfo{pages}{65--78}.
%Type = Article
\bibitem[{Koehler et~al.(2025)Koehler, Kuhm, Huffaker, Young, Tandon, Andr{\'e}, Frey, Greil, Hussain and Engelhardt}]{koehler2025strain}
\bibinfo{author}{Koehler, S.}, \bibinfo{author}{Kuhm, J.}, \bibinfo{author}{Huffaker, T.}, \bibinfo{author}{Young, D.}, \bibinfo{author}{Tandon, A.}, \bibinfo{author}{Andr{\'e}, F.}, \bibinfo{author}{Frey, N.}, \bibinfo{author}{Greil, G.}, \bibinfo{author}{Hussain, T.}, \bibinfo{author}{Engelhardt, S.}, \bibinfo{year}{2025}.
\newblock \bibinfo{title}{Deep learning--based aligned strain from cine cardiac mri for detection of fibrotic myocardial tissue in patients with duchenne muscular dystrophy}.
\newblock \bibinfo{journal}{Radiology: Artificial Intelligence} \bibinfo{volume}{7}, \bibinfo{pages}{e240303}.
%Type = Inproceedings
\bibitem[{Koehler et~al.(2022b)Koehler, Sharan, Kuhm, Ghanaat, Gordejeva, Simon, Grell, Andr{\'e} and Engelhardt}]{koehler2022comparison}
\bibinfo{author}{Koehler, S.}, \bibinfo{author}{Sharan, L.}, \bibinfo{author}{Kuhm, J.}, \bibinfo{author}{Ghanaat, A.}, \bibinfo{author}{Gordejeva, J.}, \bibinfo{author}{Simon, N.K.}, \bibinfo{author}{Grell, N.M.}, \bibinfo{author}{Andr{\'e}, F.}, \bibinfo{author}{Engelhardt, S.}, \bibinfo{year}{2022}b.
\newblock \bibinfo{title}{Comparison of evaluation metrics for landmark detection in cmr images}, in: \bibinfo{booktitle}{Bildverarbeitung f{\"u}r die Medizin 2022: Proceedings, German Workshop on Medical Image Computing, Heidelberg, June 26-28, 2022}, \bibinfo{organization}{Springer}. pp. \bibinfo{pages}{198--203}.
%Type = Inproceedings
\bibitem[{Kong et~al.(2016)Kong, Zhan, Shin, Denny and Zhang}]{phase_detection_kong2016}
\bibinfo{author}{Kong, B.}, \bibinfo{author}{Zhan, Y.}, \bibinfo{author}{Shin, M.}, \bibinfo{author}{Denny, T.}, \bibinfo{author}{Zhang, S.}, \bibinfo{year}{2016}.
\newblock \bibinfo{title}{Recognizing end-diastole and end-systole frames via deep temporal regression network}, in: \bibinfo{booktitle}{Medical Image Computing and Computer-Assisted Intervention-MICCAI 2016: 19th International Conference, Athens, Greece, October 17-21, 2016, Proceedings, Part III 19}, \bibinfo{organization}{Springer}. pp. \bibinfo{pages}{264--272}.
\newblock \DOIprefix\doi{10.1007/978-3-319-46726-9_31}.
%Type = Article
\bibitem[{Krebs et~al.(2019)Krebs, Delingette, Mailhé, Ayache and Mansi}]{cmr_probabilistiC_diffeomorphiC_registration_Krebs_2019}
\bibinfo{author}{Krebs, J.}, \bibinfo{author}{Delingette, H.}, \bibinfo{author}{Mailhé, B.}, \bibinfo{author}{Ayache, N.}, \bibinfo{author}{Mansi, T.}, \bibinfo{year}{2019}.
\newblock \bibinfo{title}{Learning a probabilistic model for diffeomorphic registration}.
\newblock \bibinfo{journal}{IEEE Transactions on Medical Imaging} \bibinfo{volume}{38}, \bibinfo{pages}{2165--2176}.
\newblock \DOIprefix\doi{10.1109/TMI.2019.2897112}.
%Type = Inproceedings
\bibitem[{Krebs et~al.(2020)Krebs, Mansi, Ayache and Delingette}]{cmr_probabilistiC_registration_Krebs_2020}
\bibinfo{author}{Krebs, J.}, \bibinfo{author}{Mansi, T.}, \bibinfo{author}{Ayache, N.}, \bibinfo{author}{Delingette, H.}, \bibinfo{year}{2020}.
\newblock \bibinfo{title}{Probabilistic motion modeling from medical image sequences: application to cardiac cine-mri}, in: \bibinfo{booktitle}{International Workshop on Statistical Atlases and Computational Models of the Heart}, \bibinfo{organization}{Springer}. pp. \bibinfo{pages}{176--185}.
\newblock \DOIprefix\doi{10.1007/978-3-030-39074-7_19}.
%Type = Article
\bibitem[{Lane et~al.(2021)Lane, Azarmehr, Jevsikov, Howard, Shun-shin, Cole, Francis and Zolgharni}]{LANE_echo2021104373}
\bibinfo{author}{Lane, E.S.}, \bibinfo{author}{Azarmehr, N.}, \bibinfo{author}{Jevsikov, J.}, \bibinfo{author}{Howard, J.P.}, \bibinfo{author}{Shun-shin, M.J.}, \bibinfo{author}{Cole, G.D.}, \bibinfo{author}{Francis, D.P.}, \bibinfo{author}{Zolgharni, M.}, \bibinfo{year}{2021}.
\newblock \bibinfo{title}{Multibeat echocardiographic phase detection using deep neural networks}.
\newblock \bibinfo{journal}{Computers in Biology and Medicine} \bibinfo{volume}{133}, \bibinfo{pages}{104373}.
\newblock \URLprefix \url{https://www.sciencedirect.com/science/article/pii/S0010482521001670}, \DOIprefix\doi{https://doi.org/10.1016/j.compbiomed.2021.104373}.
%Type = Article
\bibitem[{Mada et~al.(2015)Mada, Lysyansky, Daraban, Duchenne and Voigt}]{mada_2014_10_010}
\bibinfo{author}{Mada, R.O.}, \bibinfo{author}{Lysyansky, P.}, \bibinfo{author}{Daraban, A.M.}, \bibinfo{author}{Duchenne, J.}, \bibinfo{author}{Voigt, J.U.}, \bibinfo{year}{2015}.
\newblock \bibinfo{title}{How to define end-diastole and end-systole?}
\newblock \bibinfo{journal}{JACC: Cardiovascular Imaging} \bibinfo{volume}{8}, \bibinfo{pages}{148--157}.
\newblock \URLprefix \url{https://www.jacc.org/doi/abs/10.1016/j.jcmg.2014.10.010}, \DOIprefix\doi{10.1016/j.jcmg.2014.10.010}, \href{http://arxiv.org/abs/https://www.jacc.org/doi/pdf/10.1016/j.jcmg.2014.10.010}{{\tt arXiv:https://www.jacc.org/doi/pdf/10.1016/j.jcmg.2014.10.010}}.
%Type = Article
\bibitem[{Mart{\'\i}n-Isla et~al.(2023)Mart{\'\i}n-Isla, Campello, Izquierdo, Kushibar, Sendra-Balcells, Gkontra, Sojoudi, Fulton, Arega, Punithakumar et~al.}]{mnm2_3267857}
\bibinfo{author}{Mart{\'\i}n-Isla, C.}, \bibinfo{author}{Campello, V.M.}, \bibinfo{author}{Izquierdo, C.}, \bibinfo{author}{Kushibar, K.}, \bibinfo{author}{Sendra-Balcells, C.}, \bibinfo{author}{Gkontra, P.}, \bibinfo{author}{Sojoudi, A.}, \bibinfo{author}{Fulton, M.J.}, \bibinfo{author}{Arega, T.W.}, \bibinfo{author}{Punithakumar, K.}, et~al., \bibinfo{year}{2023}.
\newblock \bibinfo{title}{Deep learning segmentation of the right ventricle in cardiac mri: the m\&ms challenge}.
\newblock \bibinfo{journal}{IEEE Journal of Biomedical and Health Informatics} \bibinfo{volume}{27}, \bibinfo{pages}{3302--3313}.
%Type = Article
\bibitem[{Meng et~al.(2022)Meng, Qin, Bai, Liu, de~Marvao, O’Regan and Rueckert}]{cmr_motion_MulViMotion_Meng_2022}
\bibinfo{author}{Meng, Q.}, \bibinfo{author}{Qin, C.}, \bibinfo{author}{Bai, W.}, \bibinfo{author}{Liu, T.}, \bibinfo{author}{de~Marvao, A.}, \bibinfo{author}{O’Regan, D.P.}, \bibinfo{author}{Rueckert, D.}, \bibinfo{year}{2022}.
\newblock \bibinfo{title}{Mulvimotion: Shape-aware 3d myocardial motion tracking from multi-view cardiac mri}.
\newblock \bibinfo{journal}{IEEE Transactions on Medical Imaging} \bibinfo{volume}{41}, \bibinfo{pages}{1961--1974}.
\newblock \DOIprefix\doi{10.1109/TMI.2022.3154599}.
%Type = Inproceedings
\bibitem[{Qin et~al.(2018)Qin, Bai, Schlemper, Petersen, Piechnik, Neubauer and Rueckert}]{cmr_motion_estimation_Qin_2018}
\bibinfo{author}{Qin, C.}, \bibinfo{author}{Bai, W.}, \bibinfo{author}{Schlemper, J.}, \bibinfo{author}{Petersen, S.E.}, \bibinfo{author}{Piechnik, S.K.}, \bibinfo{author}{Neubauer, S.}, \bibinfo{author}{Rueckert, D.}, \bibinfo{year}{2018}.
\newblock \bibinfo{title}{Joint learning of motion estimation and segmentation for cardiac mr image sequences}, in: \bibinfo{booktitle}{Medical Image Computing and Computer Assisted Intervention--MICCAI 2018: 21st International Conference, Granada, Spain, September 16-20, 2018, Proceedings, Part II 11}, \bibinfo{organization}{Springer}. pp. \bibinfo{pages}{472--480}.
\newblock \DOIprefix\doi{10.1007/978-3-030-00934-2_53}.
%Type = Inproceedings
\bibitem[{Ronneberger et~al.(2015)Ronneberger, Fischer and Brox}]{Ronneberger_2015_U-net}
\bibinfo{author}{Ronneberger, O.}, \bibinfo{author}{Fischer, P.}, \bibinfo{author}{Brox, T.}, \bibinfo{year}{2015}.
\newblock \bibinfo{title}{U-net: Convolutional networks for biomedical image segmentation}, in: \bibinfo{editor}{Navab, N.}, \bibinfo{editor}{Hornegger, J.}, \bibinfo{editor}{Wells, W.M.}, \bibinfo{editor}{Frangi, A.F.} (Eds.), \bibinfo{booktitle}{Medical Image Computing and Computer-Assisted Intervention -- MICCAI 2015}, \bibinfo{publisher}{Springer International Publishing}, \bibinfo{address}{Cham}. pp. \bibinfo{pages}{234--241}.
%Type = Misc
\bibitem[{Sarikouch and Beerbaum(2005)}]{gcn_study_description}
\bibinfo{author}{Sarikouch, S.}, \bibinfo{author}{Beerbaum, P.}, \bibinfo{year}{2005}.
\newblock \bibinfo{title}{Follow up of post-repair tetralogy of fallot}.
\newblock \URLprefix \url{https://clinicaltrials.gov/ct2/show/NCT00266188. en}. \bibinfo{note}{last accessed 03. June, 2024}.
%Type = Article
\bibitem[{Sarikouch et~al.(2011)Sarikouch, Koerperich, Dubowy, Boethig, Boettler, Mir, Peters, Kuehne, Beerbaum and for Congenital Heart Defects~Investigators}]{gcn_sarikouch2011}
\bibinfo{author}{Sarikouch, S.}, \bibinfo{author}{Koerperich, H.}, \bibinfo{author}{Dubowy, K.O.}, \bibinfo{author}{Boethig, D.}, \bibinfo{author}{Boettler, P.}, \bibinfo{author}{Mir, T.S.}, \bibinfo{author}{Peters, B.}, \bibinfo{author}{Kuehne, T.}, \bibinfo{author}{Beerbaum, P.}, \bibinfo{author}{for Congenital Heart Defects~Investigators, G.C.N.}, \bibinfo{year}{2011}.
\newblock \bibinfo{title}{Impact of gender and age on cardiovascular function late after repair of tetralogy of fallot: percentiles based on cardiac magnetic resonance}.
\newblock \bibinfo{journal}{Circulation: Cardiovascular Imaging} \bibinfo{volume}{4}, \bibinfo{pages}{703--711}.
\newblock \DOIprefix\doi{10.1161/CIRCIMAGING.111.963637}.
%Type = Article
\bibitem[{Shalbaf et~al.(2015)Shalbaf, AlizadehSani and Behnam}]{shalbaf2015echocardiography}
\bibinfo{author}{Shalbaf, A.}, \bibinfo{author}{AlizadehSani, Z.}, \bibinfo{author}{Behnam, H.}, \bibinfo{year}{2015}.
\newblock \bibinfo{title}{Echocardiography without electrocardiogram using nonlinear dimensionality reduction methods}.
\newblock \bibinfo{journal}{Journal of Medical Ultrasonics} \bibinfo{volume}{42}, \bibinfo{pages}{137--149}.
%Type = Misc
\bibitem[{WHO(2021)}]{who_cvd_deaths19}
\bibinfo{author}{WHO, W.H.O.}, \bibinfo{year}{2021}.
\newblock \bibinfo{title}{Cardiovascular diseases (cvds)}.
\newblock \bibinfo{note}{\url{https://www.who.int/news-room/fact-sheets/detail/cardiovascular-diseases-(cvds)} (accessed: 11.02.2025)}.
%Type = Article
\bibitem[{Xue et~al.(2018)Xue, Brahm, Pandey, Leung and Li}]{xue2018cmr}
\bibinfo{author}{Xue, W.}, \bibinfo{author}{Brahm, G.}, \bibinfo{author}{Pandey, S.}, \bibinfo{author}{Leung, S.}, \bibinfo{author}{Li, S.}, \bibinfo{year}{2018}.
\newblock \bibinfo{title}{Full left ventricle quantification via deep multitask relationships learning}.
\newblock \bibinfo{journal}{Medical Image Analysis} \bibinfo{volume}{43}, \bibinfo{pages}{54--65}.
\newblock \URLprefix \url{https://www.sciencedirect.com/science/article/pii/S1361841517301366}, \DOIprefix\doi{https://doi.org/10.1016/j.media.2017.09.005}.
%Type = Article
\bibitem[{Yang et~al.(2017)Yang, He, Hussain, Xie and Lei}]{yang2017convolutional}
\bibinfo{author}{Yang, F.}, \bibinfo{author}{He, Y.}, \bibinfo{author}{Hussain, M.}, \bibinfo{author}{Xie, H.}, \bibinfo{author}{Lei, P.}, \bibinfo{year}{2017}.
\newblock \bibinfo{title}{Convolutional neural network for the detection of end-diastole and end-systole frames in free-breathing cardiac magnetic resonance imaging}.
\newblock \bibinfo{journal}{Computational and mathematical methods in medicine} \bibinfo{volume}{2017}, \bibinfo{pages}{1640835}.
%Type = Article
\bibitem[{Zolgharni et~al.(2017)Zolgharni, Negoita, Dhutia, Mielewczik, Manoharan, Sohaib, Finegold, Sacchi, Cole and Francis}]{zolgharni_2D_echo}
\bibinfo{author}{Zolgharni, M.}, \bibinfo{author}{Negoita, M.}, \bibinfo{author}{Dhutia, N.M.}, \bibinfo{author}{Mielewczik, M.}, \bibinfo{author}{Manoharan, K.}, \bibinfo{author}{Sohaib, S.A.}, \bibinfo{author}{Finegold, J.A.}, \bibinfo{author}{Sacchi, S.}, \bibinfo{author}{Cole, G.D.}, \bibinfo{author}{Francis, D.P.}, \bibinfo{year}{2017}.
\newblock \bibinfo{title}{10}.
\newblock \bibinfo{journal}{Echocardiography} \bibinfo{volume}{34}, \bibinfo{pages}{956--967}.

\end{thebibliography}

\appendix

\section{Related Work}
Detailed description of used datasets in table \ref{tbl:related_work_key-frame}.
\begin{sidewaystable*}[]
    \centering
    \tiny
    \begin{tabular}{l|l|l|c|c|c|c|c}
        \textbf{Reference}  &\textbf{Method Type} & \textbf{Labels/Inputs} & \textbf{Modality}  & \textbf{Views} & \textbf{n (Train/Eval)} & \textbf{Public} & \textbf{Cohort Type}\\
        \hline
        \cite{kachenoura2007automatic}  &Semi-Automatic & 3 Landmarks +  \acrshort{ed} frame  & Echo  &  \acrshort{2ch},  \acrshort{4ch} & 37 (–/–) & No &   Healthy   \\
        \cite{barcaro_echo}  &Semi-Automatic & Level-Set & Echo  &  \acrshort{2ch},  \acrshort{4ch} & NR  & NR &    NR \\
        \cite{gifani2010automatic}  &Unsupervised ML & None & Echo  &  \acrshort{2ch},  \acrshort{4ch} & 6 (–/–) & No &    Healthy \\
        \cite{darvishi_echo}  &Semi-Automatic & Landmark selection  & Echo  &  \acrshort{2ch},  \acrshort{4ch} & 44 (–/–) & No &   Healthy \\
        \cite{shalbaf2015echocardiography}  &Semi-Automatic & Landmark selection & Echo  &  \acrshort{2ch},  \acrshort{4ch}, \acrshort{sax} & 32 (–/–) & No &   Healthy + 2 path.  \\
        \cite{dezaki2018cardiac}  & Supervised DL & Phase labels & Echo  &  \acrshort{4ch} & 3087 (–/–) & No &  Various path.  \\
        \cite{fiorito2018detection}  &Supervised DL & Segmentation labels  & Echo  &  \acrshort{2ch},  \acrshort{4ch} & 500 (–/–) & No &  Various path. \\
        \cite{LANE_echo2021104373}  &Supervised DL & Segmentation + phase labels  & Echo  &  \acrshort{4ch} & 11070 (–/–) &  Mostly &   Various path. \\
        \cite{phase_detection_kong2016}  & Supervised DL & Phase labels  & \acrshort{cmr}  &  \acrshort{2ch},  \acrshort{4ch}, \acrshort{sax} & 420 (–/–) & No &  Various path.  \\
        \cite{xue2018cmr}  & Supervised DL & Segmentation + phase labels  & \acrshort{cmr}  & \acrshort{sax} & 145 (–/–) & No & Various path.  \\
        \cite{garcia2023cardiac}  & Supervised DL & Phase labels & \acrshort{cmr}  & \acrshort{sax} & 360 (–/–) & Yes &  Various path.  \\
        \textbf{Ours}  &\textbf{Self-Supervised} & \textbf{No additional label} & \acrshort{cmr}  &  \acrshort{4ch}, \acrshort{sax} & \textbf{200/366, 200/870}  & \textbf{Mostly} & Various path.  \\
    \end{tabular}
    \caption{Overview of keyframe/phase detection methods in cardiac imaging. “Sup.” denotes supervised methods. Public: availability of dataset(s). Reproducibility considers data access and method dependency on labels. “NR” = Not reported; “–” = Not available.}
    \label{tbl:related_work_key-frame}
\end{sidewaystable*}

\newpage
\section{Methodology}

\subsection{Datasets}
\label{sec:App-dataset}
\subsubsection{\texorpdfstring{\acrshort{mm2} \citep{mnm2_3267857}}{mm2}}
The \acrfull{mm2} dataset  \citep{mnm2_9458279, mnm2_3267857} from the 12th workshop on \acrfull{stacom} in 2021 comprises \acrshort{cmr} images with both \acrshort{sax} and \acrshort{4ch} views. It includes 360 cases of patients with seven different pathologies and healthy subjects, acquired at three Spanish clinical centres using nine different scanners from three different vendors. The dataset provides a split for training and inference. For the \acrshort{4ch} view, the test set received additional annotations of all five keyframes, while the original  \acrshort{ed} and \acrshort{es} annotation were used for \acrshort{sax} testing. 
The majority of \acrshort{cmr} sequences in both views of the test set start near the  \acrshort{ed} phase (139/140 for \acrshort{sax}/\acrshort{4ch}), with the remaining 21/20 sequences closer to the \acrshort{es} phase. For the re-labeled data in \acrshort{4ch}, a similar number of sequences start near the  \acrshort{ed} phase (145), while the remaining sequences are split between those starting near \acrshort{md} (13) and those near \acrshort{ms} (2).

\subsubsection{\texorpdfstring{\acrshort{mnms} \citep{mnm2_9458279}}{mnms}}
The \acrfull{mnms} dataset \citep{mnm2_9458279} was released as part of the \acrshort{miccai} 2020 challenge on generalisable \acrshort{cmr} segmentation. It consists of 345 short-axis \acrshort{cmr} studies from patients with various pathologies, as well as healthy subjects. These studies were collected at multiple clinical sites in Spain and Germany. Data acquisition was carried out in five hospitals using four distinct scanner vendors (Siemens, Philips, GE, Canon). 

\subsubsection{\texorpdfstring{\acrshort{acdc} \citep{acdc_dataset_8360453}}{acdc}}
The \acrfull{acdc} dataset \citep{acdc_dataset_8360453} was published as part of the \acrfull{miccai} challenge 2017. It comprises \acrshort{sax} \acrshort{cmr} from 100 patients acquired at the University Hospital of Dijon (France) using two scanners with different field strengths (1.5\,T and 3.0\,T). The dataset includes patients with four different pathologies, as well as healthy subjects. The  \acrshort{ed} phase was arbitrarily labelled frame 0 throughout the entire cohort, which is a simplification. After relabelling the original cardiac phase labels, 75 sequences start near \acrshort{ms}, while the remaining 25 sequences start close to the  \acrshort{ed} phase. Furthermore, we realised that not all 4D sequences capture an entire cardiac cycle \citep{koehler2022self}. 
% inter-observer: 
%        \acrshort{ed}: $1.13 \pm 1.64$ max 12
%       \acrshort{es}: $ 1.2 \pm 1.23 $ max 6
%       ALL: $1.16 \pm 1.45$
% \acrshort{4ch}: starting phase - re-label:
        %  \acrshort{ed}#    145
        % \acrshort{md}#     13
        % \acrshort{ms}#      2
% \acrshort{4ch}:  starting phase - original:
        %  \acrshort{ed}#    140
        % \acrshort{es}#     20  
% \acrshort{sax}: starting phase - original
        %  \acrshort{ed}#    139
        % \acrshort{es}#     21

\subsubsection{\texorpdfstring{\acrshort{gcn} \citep{gcn_sarikouch2011}}{gcn}}
For additional inference of both views the \acrfull{gcn} dataset (study identifier: NCT00266188) was employed, which was created as part of a nationwide prospective study of patients with repaired \acrfull{tof} \citep{gcn_study_description, gcn_sarikouch2011}. This dataset consists of patients with congenital heart disease (age \(17.9\pm8.3\)\,years) from 14 centres across Germany. 
% The objective was to ascertain the influence of gender and age on cardiac function by \acrshort{cmr} in patients with repaired \acrshort{tof}, which is a congenital heart disease with several dysfunctions. 
A total of 720 \acrshort{cmr} sequences in \acrshort{sax} and \acrshort{4ch}  views were recorded according to a standardized protocol from patients aged at least 8 years who had undergone \acrshort{tof} correction intervention at least one year earlier. Following the completion of the pre-processing and subsequent manual labelling by physicians, the dataset comprises 265 \acrshort{sax} \acrshort{cmr} with five keyframe labels and 206 \acrshort{4ch} \acrshort{cmr} with  \acrshort{ed} and \acrshort{es} labels. For the \acrshort{sax} view, 191 sequences start close to the \acrshort{ms} and 84 close to the  \acrshort{ed} phase, while the other three phases occurred once at the sequence start. 
\begin{figure}[ht]
    \centering
    \includegraphics[width=\linewidth]{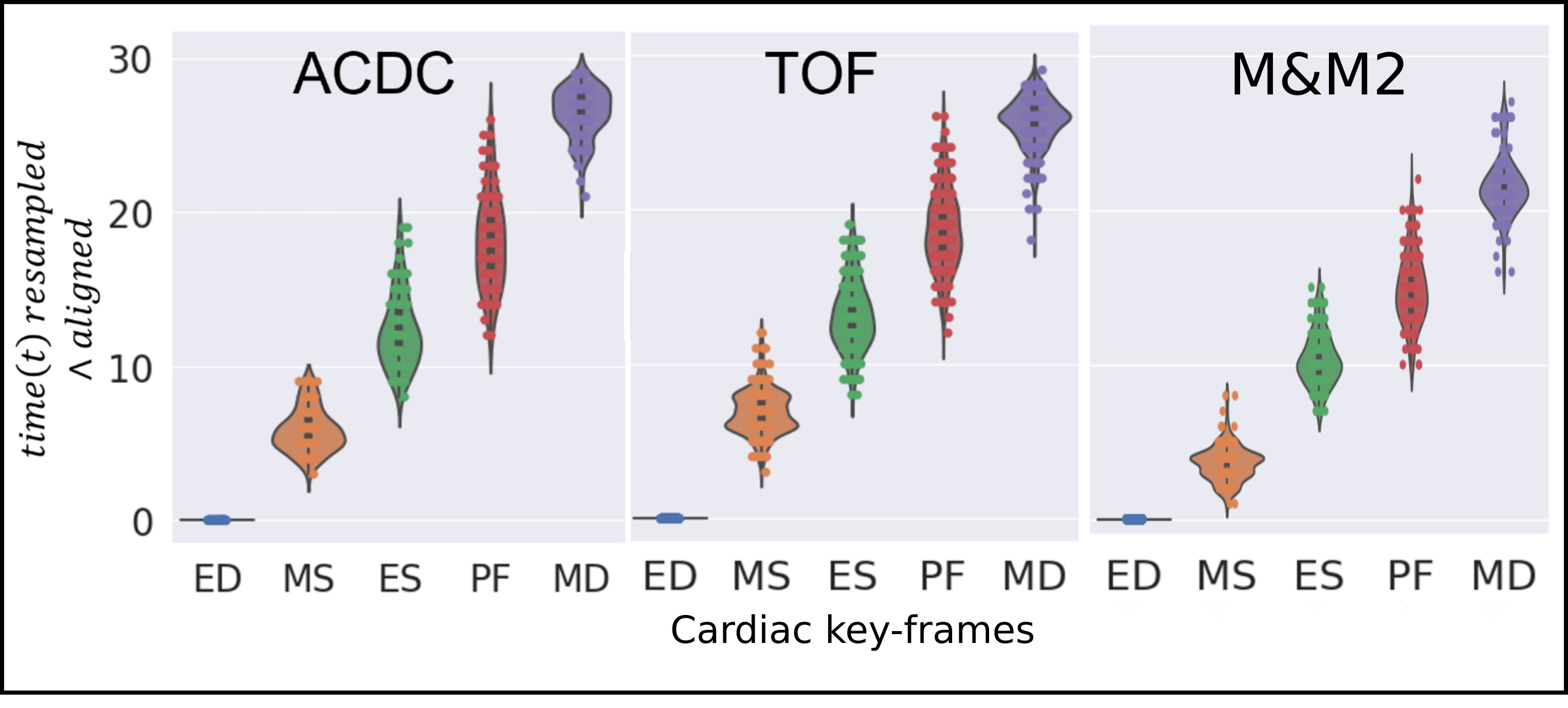}
    \caption{\textbf{Distribution of the GT phases subsequent to alignment.} In addition to the lack of alignment observed in the clinical cases, the phases demonstrate a clear overlap, which makes the comparison for physicians more complex. Distribution for \acrshort{sax} view of \acrshort{acdc} dataset and \acrshort{gcn} dataset and for  \acrshort{4ch} view of \acrshort{mm2} test dataset. }
    \label{fig:gt_phases_mm2}
\end{figure}

\newpage
\section{Results}

% \begin{table}[t]
%     \centering
%     \small
%     \begin{tabular}{c | c | c | c }
%         Dataset & Annotation &  \acrshort{ed} & \acrshort{es} \\
%         \hline
%         \acrshort{acdc} & Public vs. clinician & $1.07 \pm 0.86$ & $0.91 \pm 1.60$  \\
%         % \acrshort{acdc} & Our clinicians & & \\
%         \acrshort{mm2} & Public vs. clinician &  $1.13 \pm 1.64$ & $1.2 \pm 1.51$ 
        
%     \end{tabular}
%     \caption{Overview of inter-observer variability measured in \acrshort{cfd}}
%     \label{tab:inter_observer}
% \end{table}

\label{app2:Segmentation}
\begin{table}[t]
\centering
\small
\begin{tabular}{|c|c|c|c|c|}
\hline
\textbf{Model} & \textbf{Phase} & \textbf{Region} & \textbf{Mean ± SD} & \textbf{Median} \\ 
\hline
\acrshort{sax} & Training & \acrshort{rv} & 0.95 ± 0.03 & 0.96 \\ 
 &  & Myo & 0.86 ± 0.04 & 0.87 \\ 
 &  & \acrshort{lv} & 0.91 ± 0.06 & 0.93 \\ 
 % &  & Labels & 0.91 ± 0.03 & 0.92 \\ 
 \hline
\acrshort{sax} & Test & \acrshort{rv} & 0.94 ± 0.04 & 0.95 \\ 
 &  & Myo & 0.86 ± 0.05 & 0.87 \\ 
 &  & \acrshort{lv} & 0.90 ± 0.07 & 0.92 \\ 
 % &  & Labels & 0.91 ± 0.04 & 0.92 \\ 
\hline
\acrshort{4ch} & Training & \acrshort{rv} & 0.96 ± 0.02 & 0.96 \\ 
 & & Myo & 0.86 ± 0.08 & 0.88 \\ 
 &  & \acrshort{lv} & 0.92 ± 0.04 & 0.93 \\ 
 % &  & Labels & 0.93 ± 0.03 & 0.93 \\ 
  \hline
\acrshort{4ch} & Test & \acrshort{rv} & 0.95 ± 0.08 & 0.96 \\ 
 & & Myo & 0.84 ± 0.12 & 0.87 \\ 
 &  & \acrshort{lv} & 0.90 ± 0.10 & 0.92 \\ 
 % & & Labels & 0.91 ± 0.08 & 0.93 \\ 

\hline
\end{tabular}
\caption{DICE scores (mean ± SD and median) for \acrshort{sax} and \acrshort{4ch} segmentation models across Training and Test phases. \acrshort{rv}: Right Ventricle, Myo: Myocardium, \acrshort{lv}: Left Ventricle}
\label{tbl:segmentation_dice}
\end{table}

\begin{figure}[ht]
    \centering
    \includegraphics[width=\linewidth]{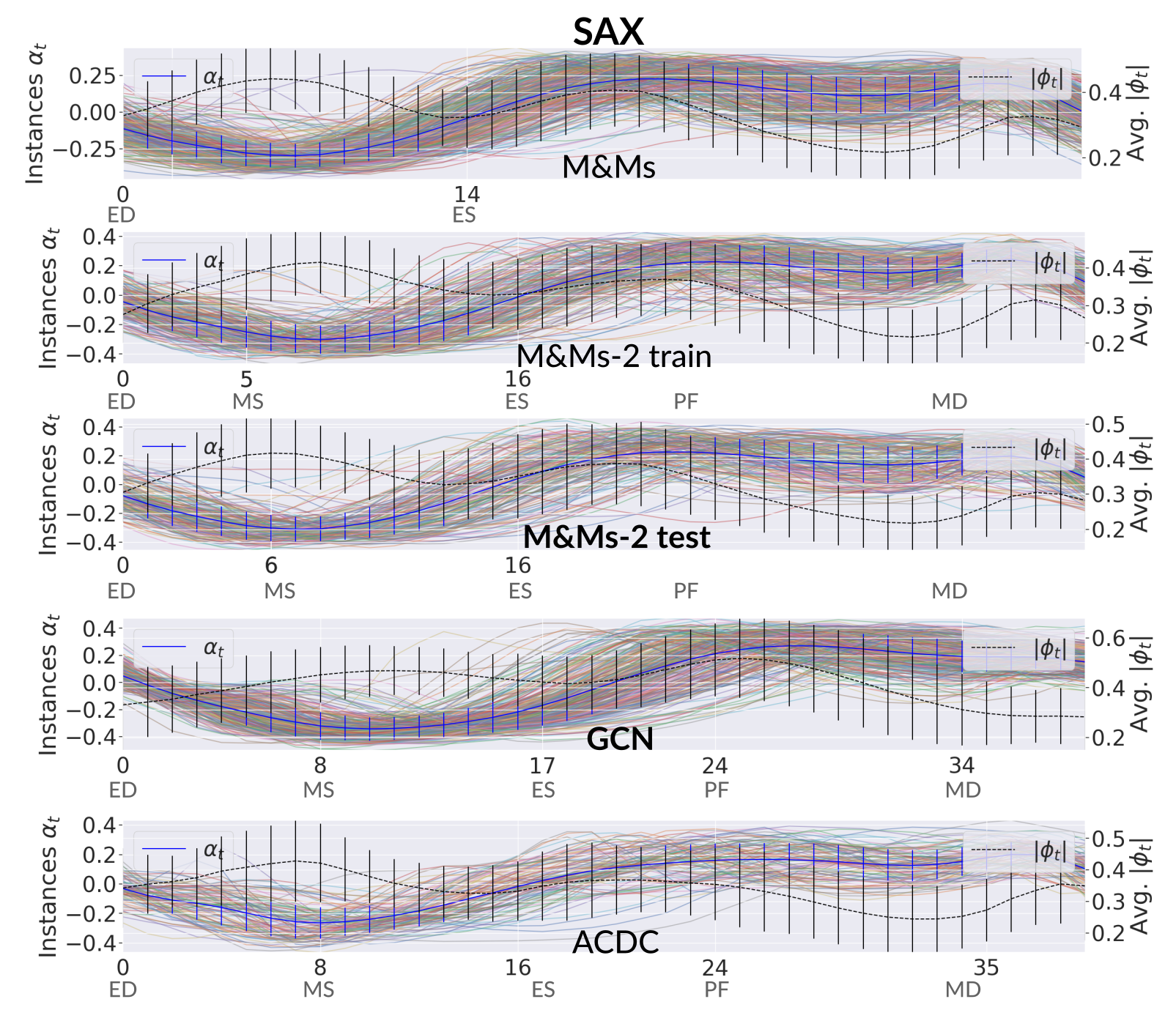}
    \caption{\textbf{Motion descriptor $\alpha_t$ for \acrshort{sax} across datasets.} Each subplot shows all instances of each dataset, linearly interpolated to 40 frames, for individual subjects represented by coloured lines. For the sake of clarity, we did not include the instance curves for $|\phi_t|$. The mean $\alpha_t$ (blue/left axis) and the $|\phi_t|$ (black/right axis) are plotted against each plot, with vertical blue and black bars representing the standard deviation respectively. The averaged phase indices (x-axis) are displayed together with the corresponding phase. In order to visualize the general properties the data was aligned at the  \acrshort{ed} phase and resized, with the original data remained unaligned.}
    \label{fig:motioncurves_SAX}
\end{figure}

\begin{figure}
    \centering
    \includegraphics[width=\linewidth]{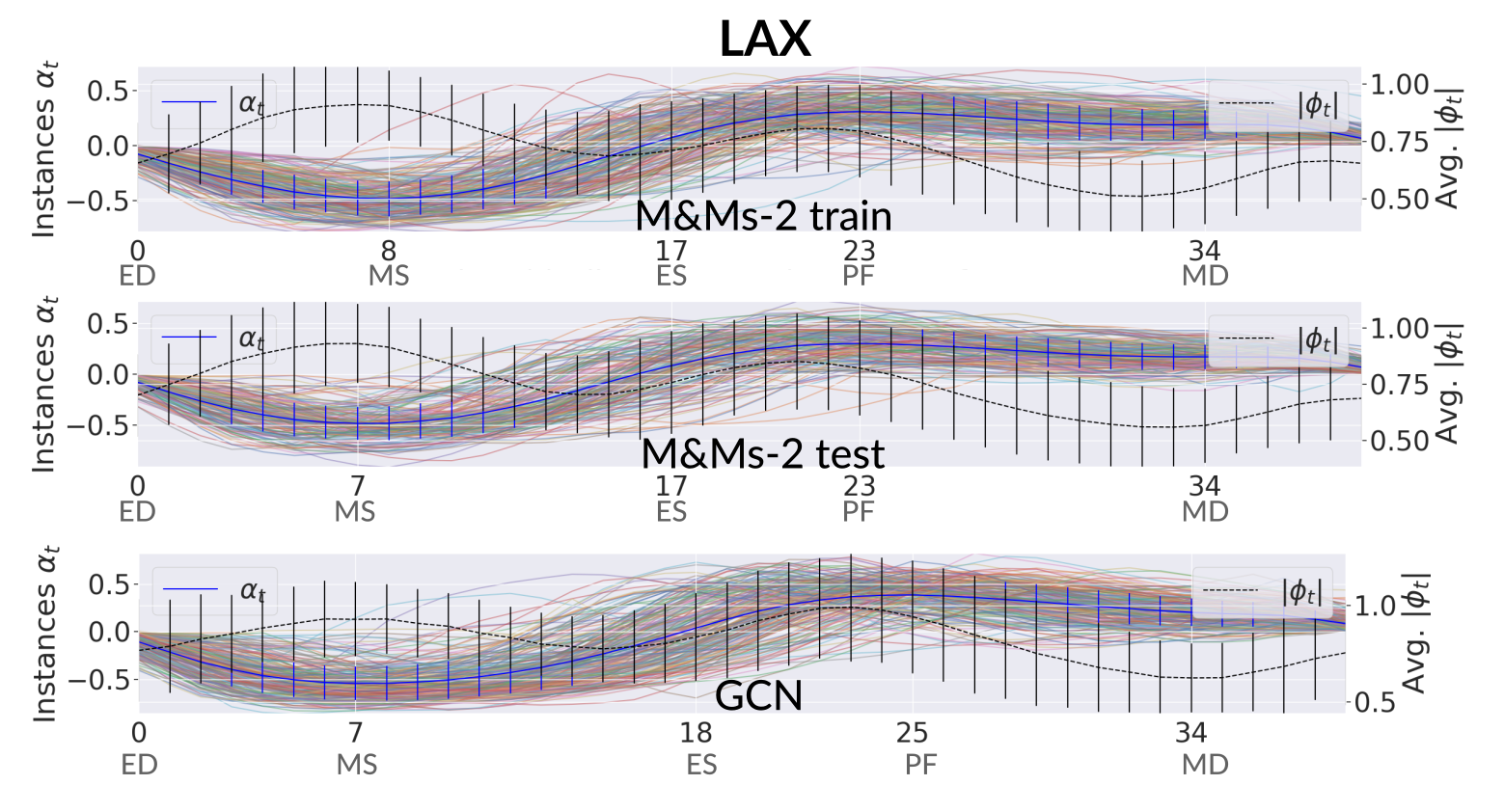}
    \caption{\textbf{Motion descriptor $\alpha_t$ for \acrshort{4ch} across datasets.} Each subplot shows all instances of each dataset, linearly interpolated to 40 frames, for individual subjects represented by coloured lines. For the sake of clarity, we did not include the instance curves for $|\phi_t|$. The mean $\alpha_t$ (blue/left axis) and the $|\phi_t|$ (black/right axis) are plotted against each plot, with vertical blue and black bars representing the standard deviation respectively. The averaged phase indices (x-axis) are displayed together with the corresponding phase. In order to visualize the general properties the data was aligned at the  \acrshort{ed} phase and resized, with the original data remained unaligned.}
    \label{fig:motioncurves_LAX}
\end{figure}

\begin{figure}[ht]
    \centering
    \includegraphics[width=0.8\linewidth]{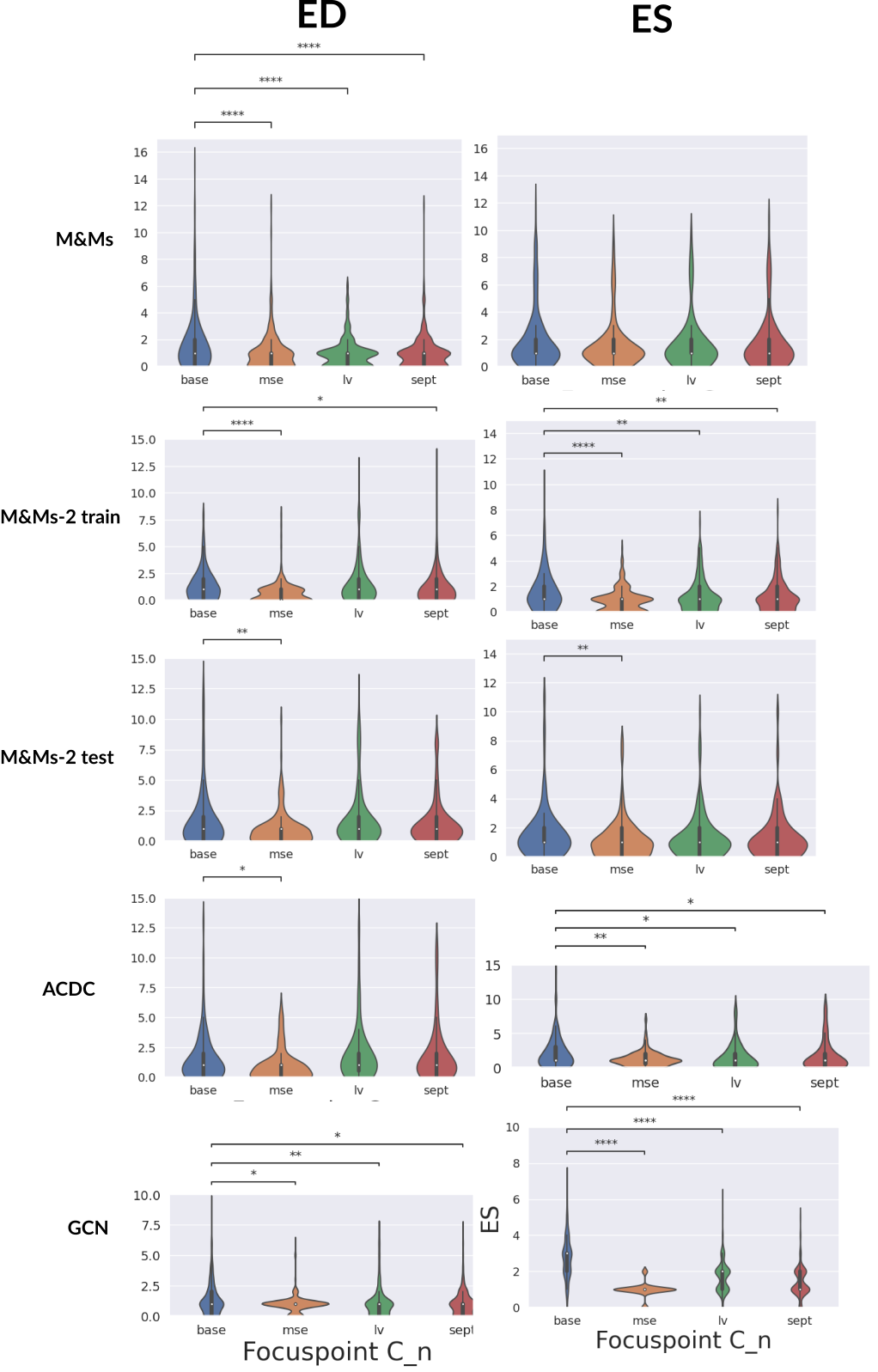}
    \caption{\textbf{Violin-plots of \acrshort{cfd} for  \acrshort{ed} and \acrshort{es} for different \acrshort{sax} datasets.} The significance per pair to the base prediction are marked with asterisk. $*: 1.00e-02 < p <= 5.00e-02$;$ **: 1.00e-03 < p <= 1.00e-02$;$ ***: 1.00e-04 < p <= 1.00e-03$; $****: p <= 1.00e-04 $}
    \label{fig:violinplts_cfd_sax}
\end{figure}

\begin{figure}[ht]
    \centering
    \includegraphics[width=0.8\linewidth]{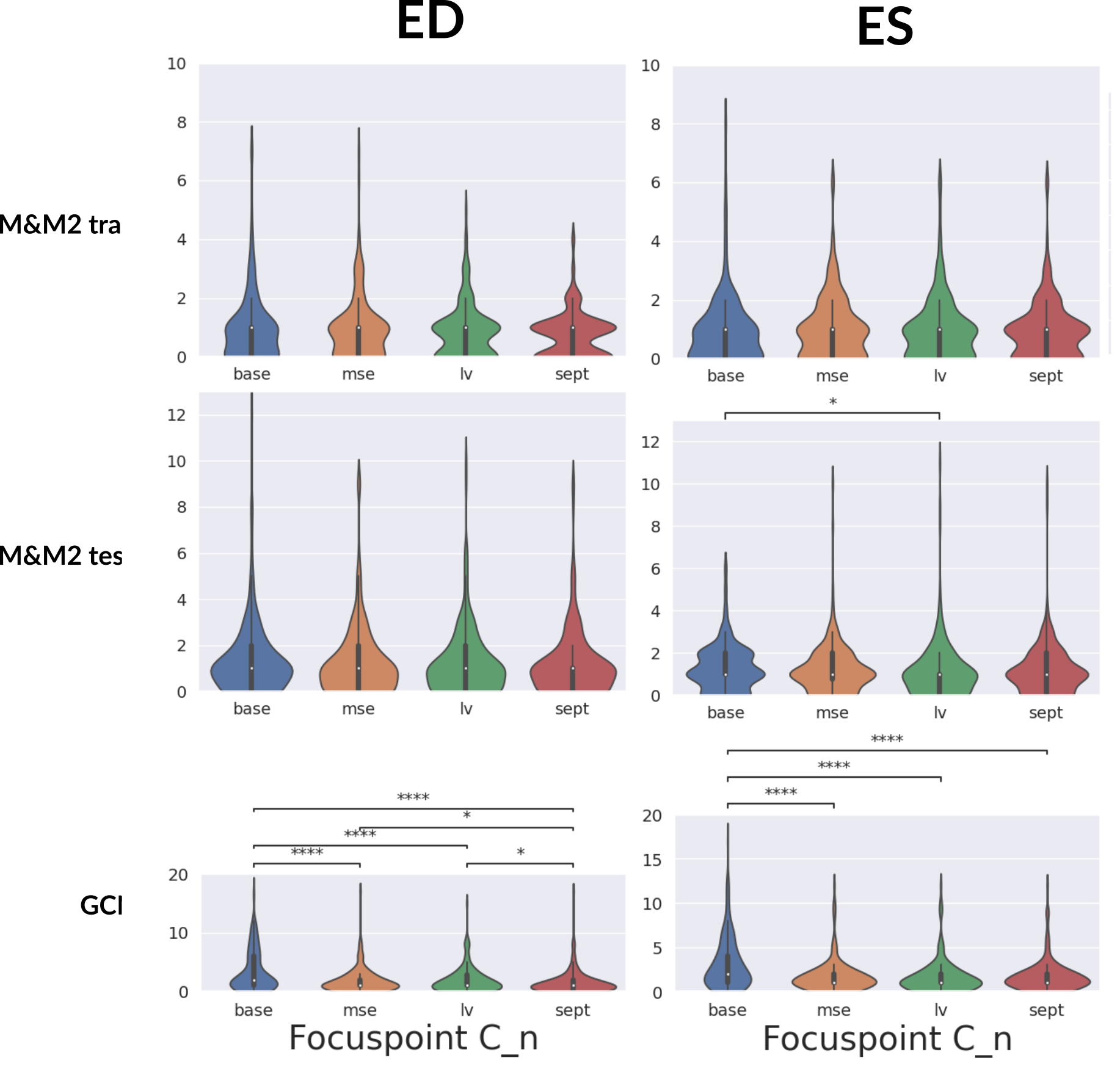}
    \caption{\textbf{Violin-plots of \acrshort{cfd} for  \acrshort{ed} and \acrshort{es} for different \acrshort{4ch} datasets.} The significance per pair to the base prediction are marked with asterisk. $*: 1.00e-02 < p <= 5.00e-02$;$ **: 1.00e-03 < p <= 1.00e-02$;$ ***: 1.00e-04 < p <= 1.00e-03$; $****: p <= 1.00e-04 $}
    \label{fig:violinplts_cfd_lax}
\end{figure}
\end{document}